\documentclass[lettersize,journal]{IEEEtran}
\usepackage{amsmath,amsfonts}
\usepackage{algorithmic}
\usepackage{algorithm}
\usepackage{array}
\usepackage[caption=false,font=normalsize,labelfont=sf,textfont=sf]{subfig}
\usepackage{textcomp}
\usepackage{stfloats}
\usepackage{url}
\usepackage{verbatim}
\usepackage{graphicx}
\usepackage{cite}

\usepackage{float}
\usepackage{booktabs}
\makeatother
\usepackage{threeparttable}
\usepackage{hyperref}
\usepackage{soul, color, xcolor}
\soulregister{\Cref}7
\soulregister{\cite}7 
\soulregister{\citep}7 
\soulregister{\citet}7 
\soulregister{\ref}7 
\soulregister{\pageref}7 
\soulregister{\frac}{2} 
\soulregister{\sum}{1} 
\soulregister{\cdot}{0} 
\soulregister{\frac}{2}    
\soulregister{\partial}{1} 
\soulregister{\sum}{1}     
\soulregister{\cdot}{0}   
\soulregister{\phi}{0}     
\soulregister{\cdots}{0}   
\soulregister{\nonumber}{0}
\usepackage{tcolorbox}
\usepackage{amsmath}
\usepackage{cleveref}
\crefname{equation}{}{} 
\crefname{figure}{fig.}{figures}
\begin{document}

\title{EMWaveNet: Physically Explainable Neural Network Based on Electromagnetic Wave Propagation for SAR Target Recognition}

\author{Zhuoxuan Li,~\IEEEmembership{Student Member,~IEEE,}~Xu Zhang,~\IEEEmembership{Graduate Student Member,~IEEE,}~Shumeng Yu, Haipeng~Wang,~\IEEEmembership{Senior Member,~IEEE}

\thanks{This work was supported in part by the National Natural Science Foundation of China (Grant No. 62271153), the Natural Science Foundation of Shanghai (Grant No. 22ZR1406700), and the Open Fund of National Key Laboratory of Scattering and Radiation, Shanghai Radio Equipment Research Institute (Grant No. 802NKL2023-002). (Corresponding author: Haipeng Wang).}
\thanks{Zhuoxuan Li, Xu Zhang, Haipeng Wang are with the Key Laboratory of Information Science of Electromagnetic Waves, Fudan University, Shanghai 200433, China (e-mail: zxli22@m.fudan.edu.cn, xuzhang19@fudan.edu.cn,  hpwang@fudan.edu.cn).}


\thanks{Shumeng Yu is with the Intelligent Medical Ultrasound Lab, Fudan University, Shanghai 200433, China (e-mail: smyu22@m.fudan.edu.cn).}

\thanks{Manuscript received xxxx, xxx; revised xxxx, xxx.}

}
\markboth{Journal of \LaTeX\ Class Files,~Vol.~14, No.~8, August~2021}%
{Shell \MakeLowercase{\textit{et al.}}: A Sample Article Using IEEEtran.cls for IEEE Journals}

\maketitle

\begin{abstract}
Deep learning technologies have significantly improved performance in the field of synthetic aperture radar (SAR) image target recognition compared to traditional methods. However, the inherent ``black box" property of deep learning models leads to a lack of transparency in decision-making processes, making them difficult to be widespread applied in practice. This is especially true in SAR applications, where the credibility and reliability of model predictions are crucial. The complexity and insufficient explainability of deep networks have become a bottleneck for their application. To tackle this issue, this study proposes a physically explainable framework for complex-valued SAR image recognition, designed based on the physical process of microwave propagation. This framework utilizes complex-valued SAR data to explore the amplitude and phase information and its intrinsic physical properties. The network architecture is fully parameterized, with all learnable parameters endowed with clear physical meanings. Experiments on both the complex-valued MSTAR dataset and a self-built Qilu-1 complex-valued dataset were conducted to validate the effectiveness of framework. The de-overlapping capability of EMWaveNet enables accurate recognition of overlapping target categories, whereas other models are nearly incapable of performing such recognition. Against 0dB forest background noise, it boasts a 20\% accuracy improvement over traditional neural networks. When targets are 60\% masked by noise, it still outperforms other models by 9\%. An end-to-end complex-valued synthetic aperture radar automatic target recognition (SAR-ATR) algorithm is constructed to perform recognition tasks in interference SAR scenarios. The results demonstrate that the proposed method possesses a strong physical decision logic, high physical explainability and robustness, as well as excellent de-aliasing capabilities. Finally, a perspective on future applications is provided.
\end{abstract}


\section{Introduction}

\IEEEPARstart{S}{ynthetic} aperture radar (SAR) represents an active microwave sensing technology capable of achieving high-resolution imagery, characterized by its all-weather, all-day imaging. Unaffected by lighting conditions or weather, SAR can provide reliable observational data even under extreme meteorological conditions. Unlike optical remote sensing images, SAR images capture the microwave characteristics of targets, with imaging results influenced by a variety of factors such as polarization modes and wavelengths, which significantly diverge from the forms of images familiar to the human visual system. This divergence leads to considerable challenges in interpretation~\cite{Oliver2004}. 
\begin{figure}[!t]
\centering
\includegraphics[width=3.7in]{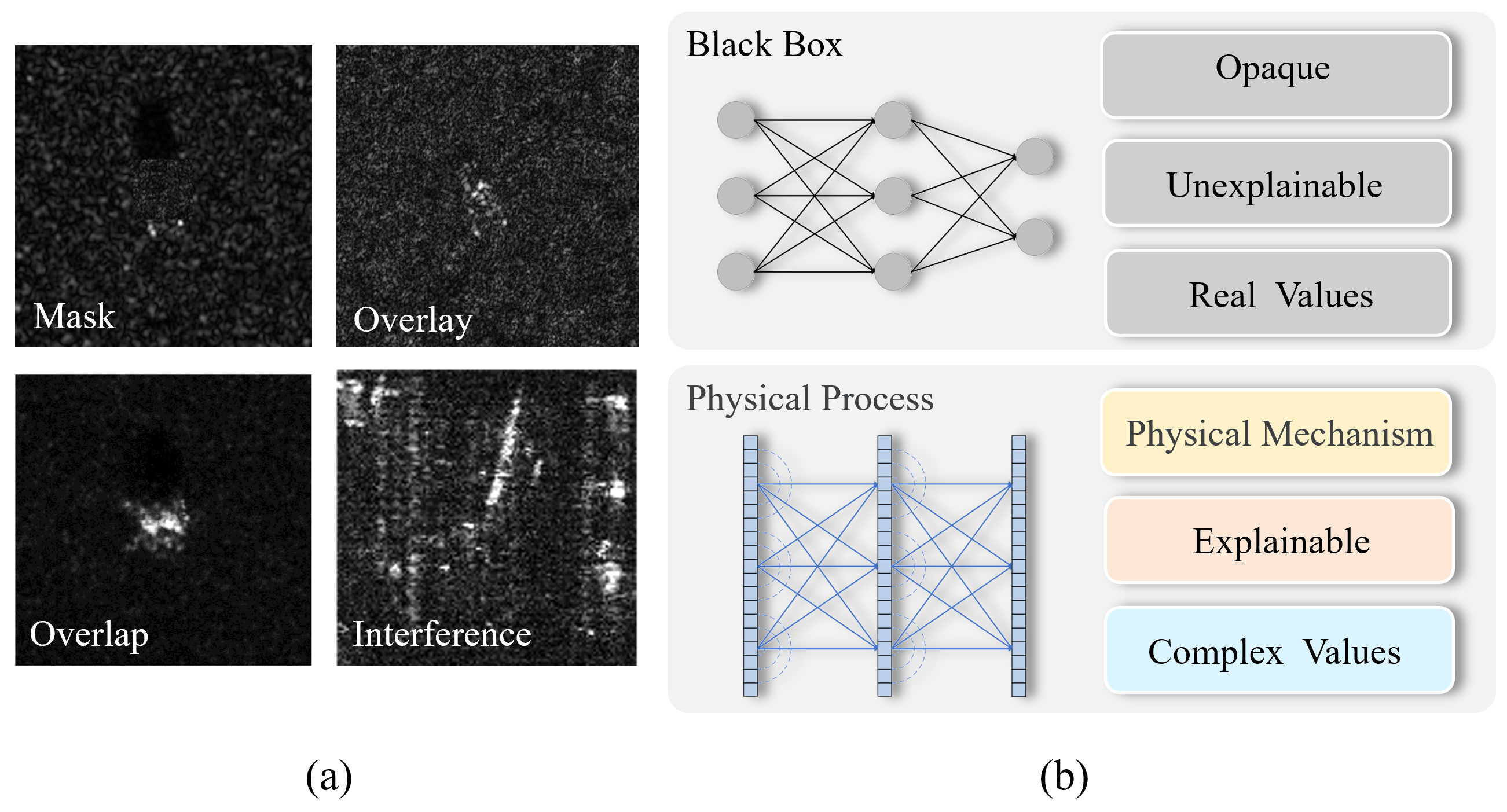}
\caption{(a) Difficulties in SAR image classification, (b) Traditional CNNs are characterized by opaque internal mechanisms and uninterpretable decision-making processes, while, our method is designed for complex-valued SAR images. The parameters learned possess clear physical significance, thereby enhancing the explainability of the network.}
\label{fig1}
\end{figure}

In recent years, the rapid development of deep learning technologies has led to increasing application in the field of SAR image interpretation~\cite{Chen2016}. Through automatic multilayer feature extraction and learning of specific tasks, deep learning has significantly outperformed traditional SAR image interpretation techniques in terms of performance. However, there exists ``hidden layer" between the input data and output results of deep learning models, with its internal processes resembling a ``black box". As a result, understanding the mechanisms behind its decisions is challenging. Explainable deep learning models enable us to understand the decision-making process. This transparency significantly reduces decision risks and minimizes the likelihood of misjudgments and errors. By analyzing the model's feedback, potential errors that adversely affect performance can be identified and corrected, thereby optimizing the model's effectiveness and enhancing the credibility and reliability of decision outcomes. Extensive research on explainability has been conducted in fields such as medicine~\cite{Patricio2022} and autonomous driving~\cite{Omeiza2022}. In the application of SAR, the lack of explainable target recognition technologies has hindered its practical deployment~\cite{Zhongling2021}. Recently, the integration of physical models has been explored in the SAR field to enhance the transparency and reliability of recognition algorithms~\cite{Weiwei2020}.

This research proposes the design of network architectures using traditional physical models that are highly explainable, to clarify the ``black box" nature of conventional neural network (CNN) models. The approach replaces opaque decision-making processes with clearly defined physical processes. To the best of our knowledge, this represents the first attempt to construct a complex-valued physics-based neural network framework for SAR target detection, wherein the network learning process is a strictly physical process and the learned parameters have clear physical meanings.

The contributions are summarized as follows
\begin{itemize}
\item[1)] A physically explainable neural network architecture is introduced founded on the propagation of electromagnetic waves in passive space for SAR target recognition. The network is fully parameterized with each parameter bearing a clearly physical significance, the learning and decision-making process of the network is a strictly physical process.

\item[2)] A novel loss function is developed named signal to noise ratio loss \(L_{SNR}\) for the network. This function classifies input targets by comparing the energy ratios between designated regions and the other detection layer. All network parameters are complex-valued and calculated in the frequency domain, effectively utilizing the phase information inherent in SAR images.

\item[3)] The network is validated on MSTAR dataset and self-built Qilu-1 dataset. Experiments were conducted under conditions of target overlap, noise interference, noise overlay, and random masking, with the results being thoroughly analyzed and discussed. This demonstrated an enhancement in the robustness and physical explainability.
\end{itemize}

The organization of this article is given as follows. The current state of explainable deep learning methods and their application in the field of SAR image interpretation are introduced in Section II. The proposed EMWaveNet is detailed in Section III. Experimental results and discussions are presented in Section IV. Finally, conclusions are drawn in Section V.

\section{Related Work}
\subsection{Explainable Deep Learning}

The explainability is always an Achilles’ heel of CNNs, and has presented challenges for decades~\cite{Zhang2018}. Selvaraju \textit{et al}.~\cite{Selvaraju2017} introduced the Gradient-weighted class activation mapping (Grad-CAM), a technique that uses gradient weights to produce class activation maps, highlighting image regions crucial for the model’s decision-making. Wang \textit{et al}.~\cite{Xiong2022} employed causal intervention to replace traditional likelihood methods, aiming to eliminate spurious correlations and enhance the model’s ability to generalize accurately. Local interpretable model-agnostic explanations (LIME) proposed by Mishra \textit{et al}. ~\cite{Mishra2017} constructs a local interpretable model near the prediction of interest to approximate the behavior of a complex model. It is model-agnostic, applicable to any model, and provides post-hoc explanations. The shapley additive explanations (SHAP) algorithm~\cite{Ribeiro2016} employed game theory to attribute the model’s output to individual pixels, revealing each pixel’s impact on the overall decision process. Deconvolution is used in neural networks to increase feature map resolution, improving visualization and enabling more precise reconstruction of input images for better explainability~\cite{Dosovitskiy2016}.

 The mathematical essence of the attention model is a weighting strategy for data,  which extracts global features compared with the small receptive field area of the CNNs. The attention matrix reflects the characteristics of interest in the decision-making process of the model, making the actions of the model easier to understand. By visualizing attention weight matrices, it becomes possible to pinpoint the specific areas of an image that are relevant to each generated result. The transformer~\cite{Vaswani2017},~\cite{Liu2021a},~\cite{Dosovitskiy2020}, a model entirely constituted by self-attention mechanisms, surpasses CNNs in terms of explainability.  
 
Current AI explainability methods are being applied to enhance model clarity in the SAR field. Visualization techniques (Grad-Cam, LIME, SHAP) link model outputs directly to specific SAR image features to illuminate the decision-making process~\cite{panati2022feature}. Additionally, the integration of Bayesian networks~\cite{Huang2023a} for uncertainty quantification and knowledge graphs~\cite{Qian2024} for structured knowledge representation are employed to augment explainability. The adoption of Transformers~\cite{Zhang2022},~\cite{Deng2021} in SAR target recognition improves performance. These methods enhance the interpretability of deep models in the SAR domain, but lack the incorporation of SAR physical information.

\subsection{Physically Explainable Deep Learning for SAR}

For SAR imagery, analyses of physical scattering properties can be based on well-founded physical models, such as polarization decomposition models for fully polarized SAR images~\cite{Ji2015}, time-frequency analysis model, and models describing the scattering center properties of targets~\cite{Potter1997}. These physical models inherently provide explainable feature representations, meaningful priors, and reduce the parameters that networks need to learn~\cite{Huang2023}. Huang \textit{et al}.~\cite{Huang2020} introduced deep SAR net (DSN), a method for SAR image classification that fuses features in both time and frequency domains. Li \textit{et al}.~\cite{Li2022} developed a method for SAR target recognition that combines global and component information within a deep convolutional neural network framework, enhancing the physical explainability of target features and thus improving model classification accuracy. Zhang \textit{et al}.~\cite{Zhang2020}, Feng \textit{et al}.~\cite{Feng2022},~\cite{Feng2021} and Liu \textit{et al}.~\cite{Liu2021} achieved robust performance by integrating scattering information extracted using attribute scattering center (ASC) with relevant information from convolutional neural networks. Huang \textit{et al}.~\cite{Huang2022} proposed a novel unsupervised approach for infusing physical guidance into deep convolutional neural networks, enhancing the explainability of deep learning models. Chan \textit{et al}.~\cite{Chan2015} introduced PCANet, an unsupervised deep convolutional neural network for image classification, where cascaded principal component analysis is utilized to construct filters. Li \textit{et al}.~\cite{Li2019} applied PCANet to the SAR images, increasing the explainability of model. Liang \textit{et al}.~\cite{Liang2020} proposed combining the iterative shrinkage thresholding algorithm with deep convolutional neural networks, enabling the model to automatically learn optimal parameters and thus possess physical explainability. The exploration of uncertainty has been proposed to enhance the explainability of SAR target recognition.
\begin{figure*}[ht]
\centering
\includegraphics[width=6.5in]{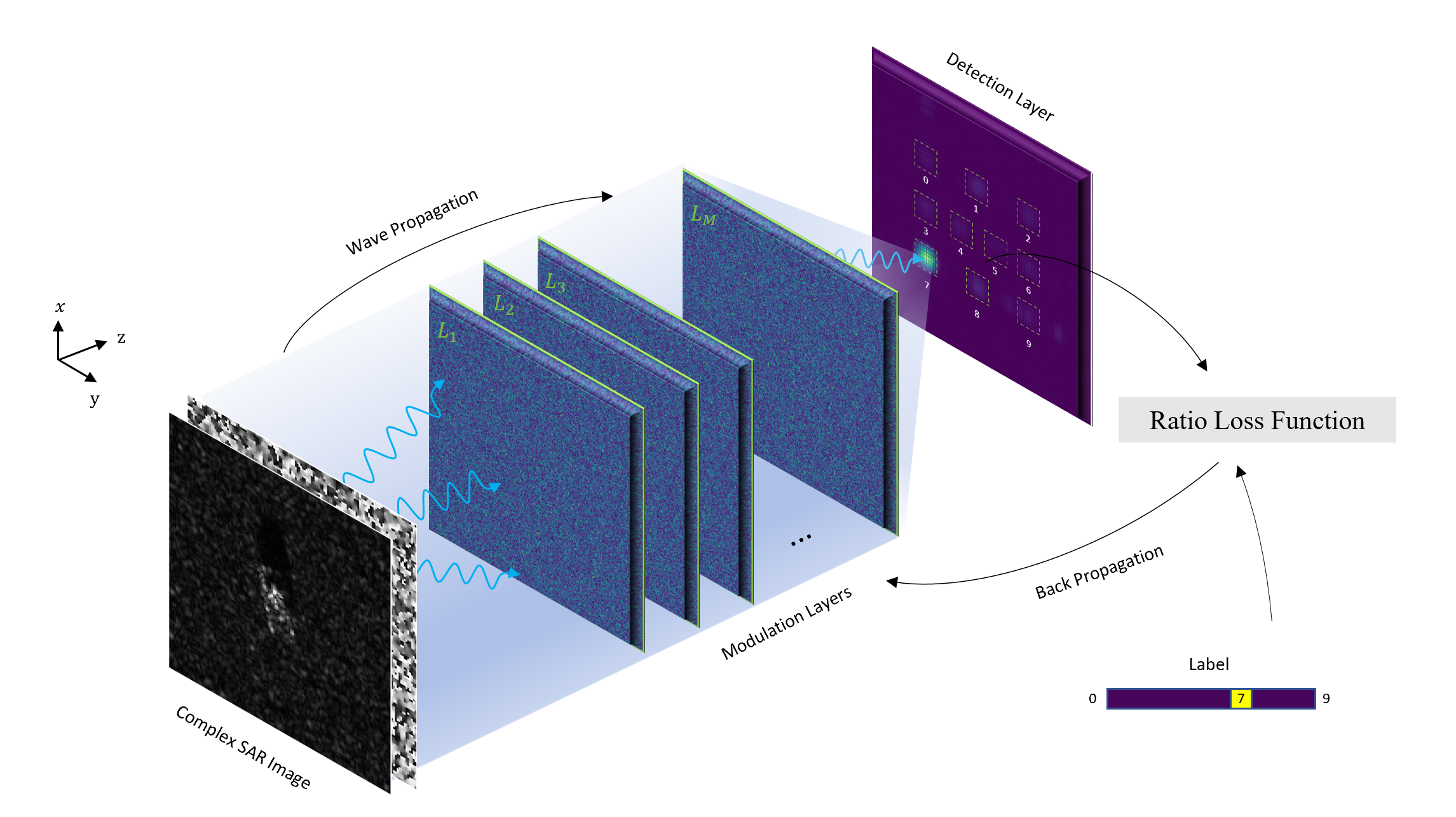}
\caption{The overall of EMWaveNet. The network consists of three parts. The input complex-valued SAR image, several modulation layers in the middle and the final
classification detection layer.}
\label{fig222}
\end{figure*}

\subsection{Complex-valued Physically Explainable Framework for SAR}
Traditional deep learning models are designed primarily for three-channel RGB optical images and are not well-suited for complex-valued SAR imagery. Many scholars have previously researched how to construct complex-valued deep learning networks to more effectively extract features from complex-valued SAR images~\cite{Trabelsi2017}. Zhang \textit{et al}.~\cite{Zhang2017} proposed a complex-valued Convolutional Neural Network (CNN) for SAR imagery, where all computations are performed in the complex-valued domain and complex-valued label is designed to fit the task, achieving significant results. Wilmanski \textit{et al}.~\cite{Wilmanski2016} developed neural networks for SAR target recognition, utilizing a complex-valued first convolution layer and real-valued subsequent layers, with an activation function outputting the magnitude of input feature map. However, to enable the network to learn more physical information, it is essential to incorporate physical models in addition to utilizing complex SAR images. Current research predominantly focuses on the fusion of physical features without embedding physical models into the design of the model framework.

\begin{figure*}[ht]
\centering
\includegraphics[width=5.5in]{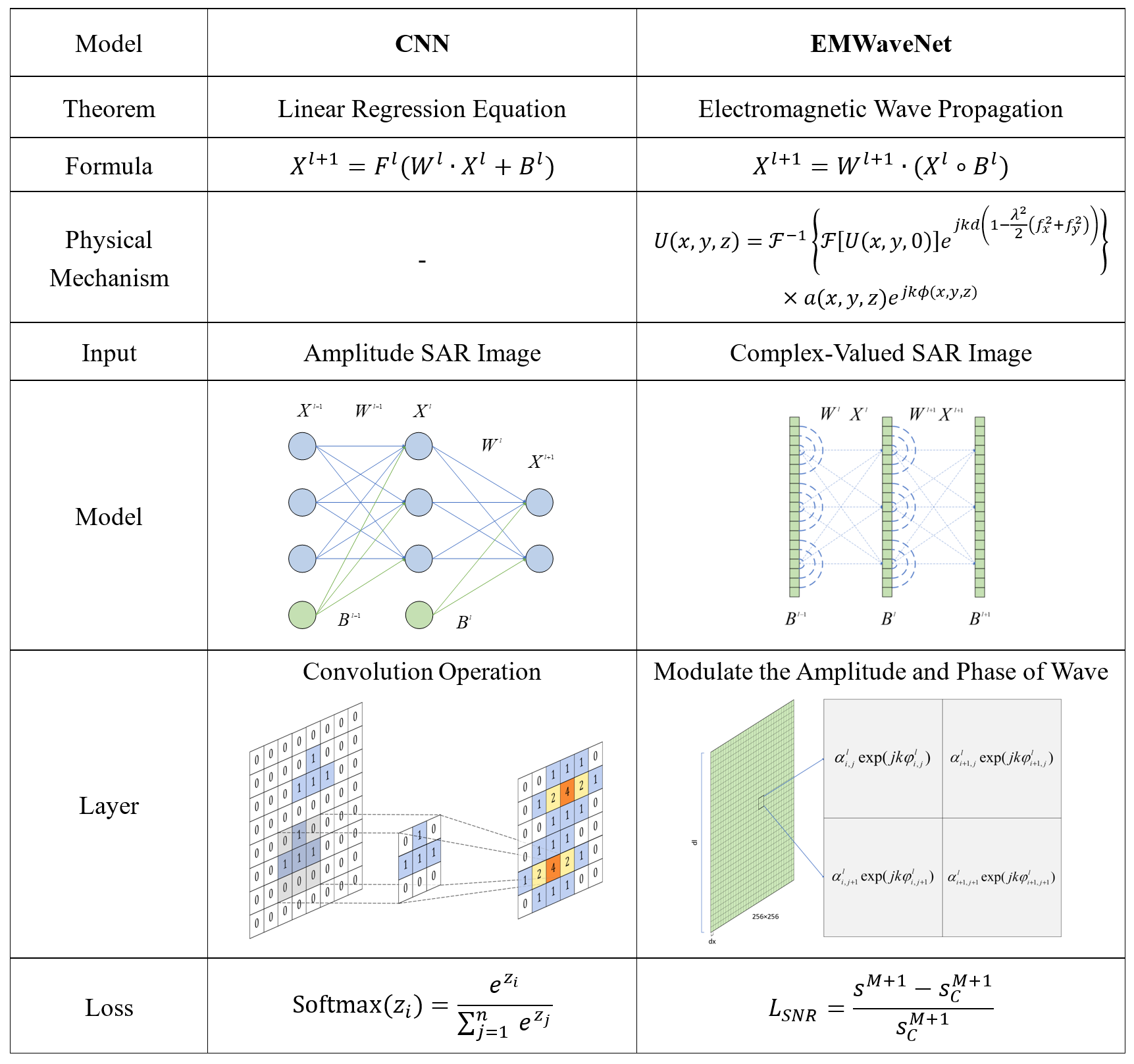}
\caption{CNN vs. EMWaveNet}
\label{fig_com}
\end{figure*}

\begin{table*}[ht]
\caption{\textbf{Compositional Structure of the Proposed EMWaveNet}}
\centering
\begin{tabular}{ccc}
\toprule
Components&Definition&Caption\\
\midrule
Input Layer &Complex-Valued SAR image & -  \\
Modulation Layer & Modulate the amplitude and phase of electromagnetic waves & Fig. 5  \\
Classification Layer & Classification is achieved by calculating the regional energy corresponding to each category &  Fig. 6  \\
\bottomrule
\label{tcom}
\end{tabular}
\end{table*}

 Research on creating an entirely new explainable framework has already been conducted in the optical field.
 Zhang \textit{et al}.~\cite{Zhang2018a} addressed the issue of convolutional kernel aliasing by grouping kernels, altering the network structure to improve model explainability. They also employ decision trees to construct the decision-making process of convolutional neural networks, thereby making the logic of network transparent~\cite{Zhang2019}. Yu \textit{et al}.~\cite{Yu2020} introduced MCR2, a method for measuring the size of feature spaces using encoding length, providing a quantitative and mathematically derivable measure for data separability. Building on this method, they derived white-box CNNs~\cite{Chan2022} and Transformers~\cite{Yu2024}, proposing an explainable and stable closed-loop system~\cite{Dai2023}. The D2NN introduced by Lin \textit{et al}.~\cite{Lin2018} is an all-optical network capable of classifying MNIST images in the THz band, and can process tasks in parallel~\cite{huang2022orbital}. 
 
 Inspired by these methods and based on the physical imaging mechanism of SAR images, this research designed a complex-valued physically explainable deep learning framework, where all parameters possess clear physical significance.

\section{EMWaveNet}

A novel physically explainable deep learning network named EMWaveNet is proposed in this section, leveraging electromagnetic propagation for complex-valued SAR image recognition. The entire framework is illustrated in \Cref{fig222}. EMWaveNet consists of an input layer, an output layer, and several intermediate microwave modulation layers, with detailed descriptions provided in \Cref{tcom}. Additionally, \Cref{fig_com} describes the differences between the proposed EMWaveNet and traditional CNN architectures.

The recognition process of the EMWaveNet interprets the complex SAR image as a complex wave field. As the wave propagates through a vacuum, the learnable modulation layers adjust the wave to concentrate its energy within specific regions of the classification layer corresponding to target categories.

\subsection{Theoretical Foundation}

In vacuum and source-free conditions, electromagnetic field perturbations under steady-state or frequency-domain analysis can be described by the Helmholtz equation

\begin{equation}
\label{e4}
\nabla^2U+k^2U=0
\end{equation}
where \(U\) represents the electromagnetic field. \(\nabla^2\)=\(\frac{\partial^2}{\partial x^2}+\frac{\partial^2}{\partial y^2}+\frac{\partial^2}{\partial z^2}\) is the Laplace operator in three-dimensional space, a differential operator. \(k\) denotes the wave number, related to the wavelength and frequency of the wave. The wave number is defined as \(k = \frac{2\pi}{\lambda}\), with \(\lambda\) being the wavelength. 


The \Cref{e4} can be solved using integral transform methods, by applying a Fourier transform to \(U=U(x, y, z)\) on the xy-plane, which correlates the spatial coordinates \((x, y)\) with the frequency domain \((f_x, f_y)\). The result after two-dimensional Fast Fourier Transform (2D-FFT) is as follows

\begin{equation}
\label{e7}
-4\pi^2(f_x^2+f_y^2)G_z+\frac{\mathrm{d}^2G_z}{\mathrm{d}z^2}+k^2G_z=0.
\end{equation}
where \(G_z(f_x, f_y, z)\) represents the 2D-FFT result of  \(U(x, y, z)\) over the \(x\) and \(y\) dimensions. Rearranging \Cref{e7} yields the following expression

\begin{equation}
\label{e8}
\frac{\mathrm{d}^2G_z}{\mathrm{d}z^2}+\left[\frac{2\pi}\lambda\sqrt{1-(\lambda f_x)^2-(\lambda f_y)^2}\right]^2G_z=0.
\end{equation}

The solution is given by

\begin{equation}
\label{e9}
G_z(f_x,f_y,z)=G_z(f_x,f_y,0)e^{i\frac{2\pi}\lambda\sqrt{1-(\lambda f_x)^2-(\lambda f_y)^2}z}.
\end{equation}
The relationship between the electromagnetic field at various points along the z-axis is established by \Cref{e9}. The electromagnetic field at \((x, y, z)\) can be obtained

\begin{equation}
\label{e10}
U(x,y,z)=\mathcal{F}^{-1}\left\{\mathcal{F}[U(x,y,0)]e^{i\frac{2\pi}\lambda\sqrt{1-(\lambda f_X)^2-(\lambda f_Y)^2}z}\right\}.
\end{equation}

The transfer function can be written as

\begin{equation}
\label{e1111}
H(f_x,f_y)=\exp{[jkd(1-\frac{\lambda^2}2(f_x^2+f_y^2))]}.
\end{equation}
where \(d\) represents the distance of propagation. Expanding \Cref{e1111} yields

\begin{equation}
H(f_x,f_y)=\exp{(j\frac{2\pi}\lambda z)}\mathrm{exp}{(-j\lambda\pi d\phi_f)}.
\label{e144}
\end{equation}
where \(\phi_f\) denotes \((f_x^2+f_y^2)\). As proven by Sherman \textit{et al}.~\cite{Sherman1967}, the inverse Fourier transform of 
\(H(f_x,f_y)\) is equivalent to the first Rayleigh-Sommerfeld solution 

\begin{align}
h(x,y,z)&=\mathcal{F}^{-1}\{H(f_x,f_y)\} \nonumber \\
&=\frac{z-z_{i}}{r^{2}}\left(\frac{1}{2\pi r}+\frac{1}{j\lambda}\right)\exp\left(\frac{j2\pi r}{\lambda}\right).
\label{e16}
\end{align}

The neural network's first convolutional layer is pivotal in learning linear structures and edge features from input images~\cite{Zeiler2011}. Notably, in AlexNet, kernels within initial layer exhibit similarities to Gabor filter~\cite{Luan2018}. Two-dimensional Gabor filter is defined by the product of a plane wave and a Gaussian function:

\begin{align}
G(x, y) &= \exp \left(-\frac{x_{0}^{2}+\gamma^{2} y_{0}^{2}}{2 \sigma^{2}}\right) \times \exp \left(j \frac{2 \pi x_{0}}{\lambda}\right) \label{eq:main} \\
x_{0} &= x \cos \theta + y \sin \theta \notag \\
y_{0} &= x \sin \theta + y \cos \theta \notag
\label{eqq}
\end{align}
where \(x\) and \(y\)  represent the pixels in the space domain of an image, \(\theta\) represents the orientation of the stripes, \(\sigma^{2}\) is the standard deviation of the Gaussian envelope, \(\gamma\) represents the spatial aspect ratio.


\begin{figure}[bp]
\centering
\includegraphics[width=3.5in]{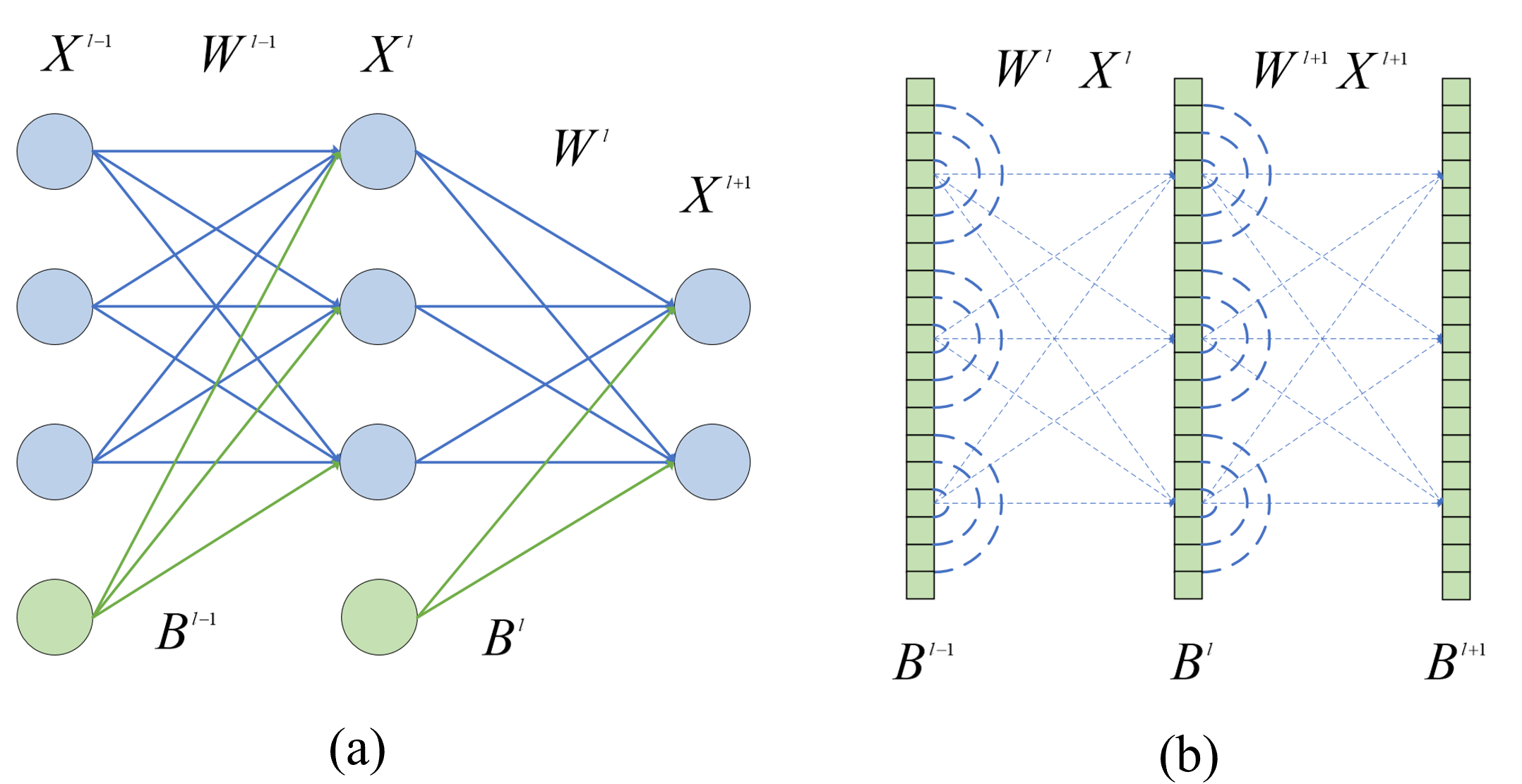}
\caption{Comparison of two network structures. (a) Traditional Convolutional Neural Network: The architecture consists of layers of neurons where each neuron in a given layer is connected to the neurons in the previous and subsequent layers. These connections allow the network to learn spatial hierarchies of features through convolution and pooling operations. (b) EMWaveNet: This network is designed to model the propagation of electromagnetic waves in a vacuum, where the process of wave propagation is modulated by the network’s parameters. EMWaveNet incorporates the principles of electromagnetic wave theory to enhance the network’s ability to capture complex wave behavior in SAR imaging.}
\label{fig3}
\end{figure}

Inspired by network visualization theory and the Gabor filter, a comparison between \Cref{e16} and (9) reveals that the propagation formulas for electromagnetic waves in a passive space resemble the Gabor filter. This suggests that the propagation process can also be regarded as a filtering mechanism. This correlation is explicitly discussed in~\cite{Goodman1968}. Thus, the process of the propagation can be regarded as a linear spatial filter capable of extracting specific features from images. This serves as the theoretical foundation for employing the propagation of electromagnetic waves in image classification. Combined with the complex-valued characteristics of SAR images and the filtering properties of the transfer function, this study constructs a complex-valued physical neural network based on the transfer function as indicated in \Cref{e1111}, which would guide the network integrates the amplitude and phase information into its architecture and processing.

\subsection{Forward Propagation}

The forward propagation formula of the traditional neural network can be written as

\begin{equation}
X^{l+1} = F^l(W^l\cdot X^l + B^l)
\label{e17}
\end{equation}
where \(l\) indicates the \(l\)-th layer within the network, the function \(F\) stands for the activation function, \(W\) for the weight matrix, \(X\) for the input data, and \(B\) for the bias. 

Similarly, the forward propagation formula of EMWaveNet can be expressed as

\begin{equation}
X^{l+1} = W^{l+1}\cdot (X^l \circ B^l).
\label{e18}
\end{equation}
where the symbol \(\circ\) denotes the Hadamard product, representing element-wise multiplication. A comparison of \Cref{e17} and \Cref{e18} reveals that both networks have two parameters: \(W\) and \(B\). However, in conventional neural networks, the bias is added to the weighted sum of data from the preceding layer, whereas in EMWaveNet, the bias is directly multiplied with the input data.

Unlike traditional networks, the weights and biases in the proposed EMWaveNet have clear physical interpretation. The physical expressions are given by (12). 

\begin{equation}
\begin{split}
U(x,y,z) = &\mathcal{F}^{-1} \{ \mathcal{F}[U(x,y,0)]e^{jkd(1-\frac{\lambda^{2}}{2}(f_{x}^{2}+f_{y}^{2}))} \} \\
&\times a(x,y,z) e^{jk\phi(x,y,z)}.
\label{e23}
\end{split}
\end{equation}
where \(W\)\(=\)\(\mathcal{F}^{-1}\{\mathcal{F}[U(x,y,0)]e^{jkd(1-\frac{\lambda^{2}}{2}(f_{x}^{2}+f_{y}^{2}))}\)\}, represents the transmission function of electromagnetic waves in a vacuum. While \(B\)\(=\)\(a(x,y,z) e^{jk\phi(x,y,z)}\), with \(a\) and \(\phi\) as the two learnable parameters that modulate the amplitude and phase of the waves. Based on \Cref{e23}, the propagation of microwaves through the various modulation layers of the network, as well as the parameter learning process can be simulated.

As illustrated in \Cref{fig3}, the network architectures of the traditional neural network and the proposed EMWaveNet are compared. Although both networks share the parameters \(W\) and \(B\), the traditional neural network learns through convolutional operations, while the proposed network learns layer by layer based on the propagation of electromagnetic waves in space.

Since SAR images are radar images with wave-like properties, a complex-valued SAR image can be regarded as a microwave wavefield. Modulation layers are used to modulate the microwaves originating from the SAR image. Each pixel within the microwave modulation layer can be considered a modulating neuron. These neurons are capable of modulating the amplitude and phase of the wave. The \(n_i^l(x,y,z)\) can be used to represent the \(i\)-th neuron in modulation layer \(l\). It can be calculated by the following


\begin{equation}
\begin{split}
\label{deqn_ex1a}
n_i^l(x,y,z) &= h_i^l(x,y,z) \cdot t_i^l(x_i,y_i,z_i) \cdot \sum_k n_k^{l-1}(x_i,y_i,z_i) \\
\end{split}
\end{equation}
where \(t_i^l=a_i^l\exp{(jk\phi_i^l)}\), represents the transmission coefficient, acting as a bias. \( h \) denotes the time-domain form of the transfer function. 

The forward propagation between layers of EMWaveNet can be represented as

\begin{equation}
n_{i,p}^l=h_{i,p}^l\cdot t_i^l\cdot m_i^l
\label{eqn:traditionalNN}
\end{equation}
\vspace{-1em} 
\begin{equation}
m_i^l=\sum_q^ln_{q,i}^{l-1}
\label{eqn:MWaveNet}
\end{equation}
where \( n \) denotes the aforementioned propagation formula for computing the transmission to the next layer. \(q\) refers to ~\(q\)-th neuron in
modulation layer \(l-1\), \(p\) refers to \(p\)-th neuron in
modulation layer \(l+1\). \( m \) represents the weighted sum at a certain node in this layer from all the neuronal nodes in the previous layer.

To summarize, the forward propagation process of EMWaveNet is based on the propagation and modulation of the microwave wavefield of a complex-valued SAR image in a vacuum.

\begin{figure}[!h]
\centering
\includegraphics[width=2.7in]{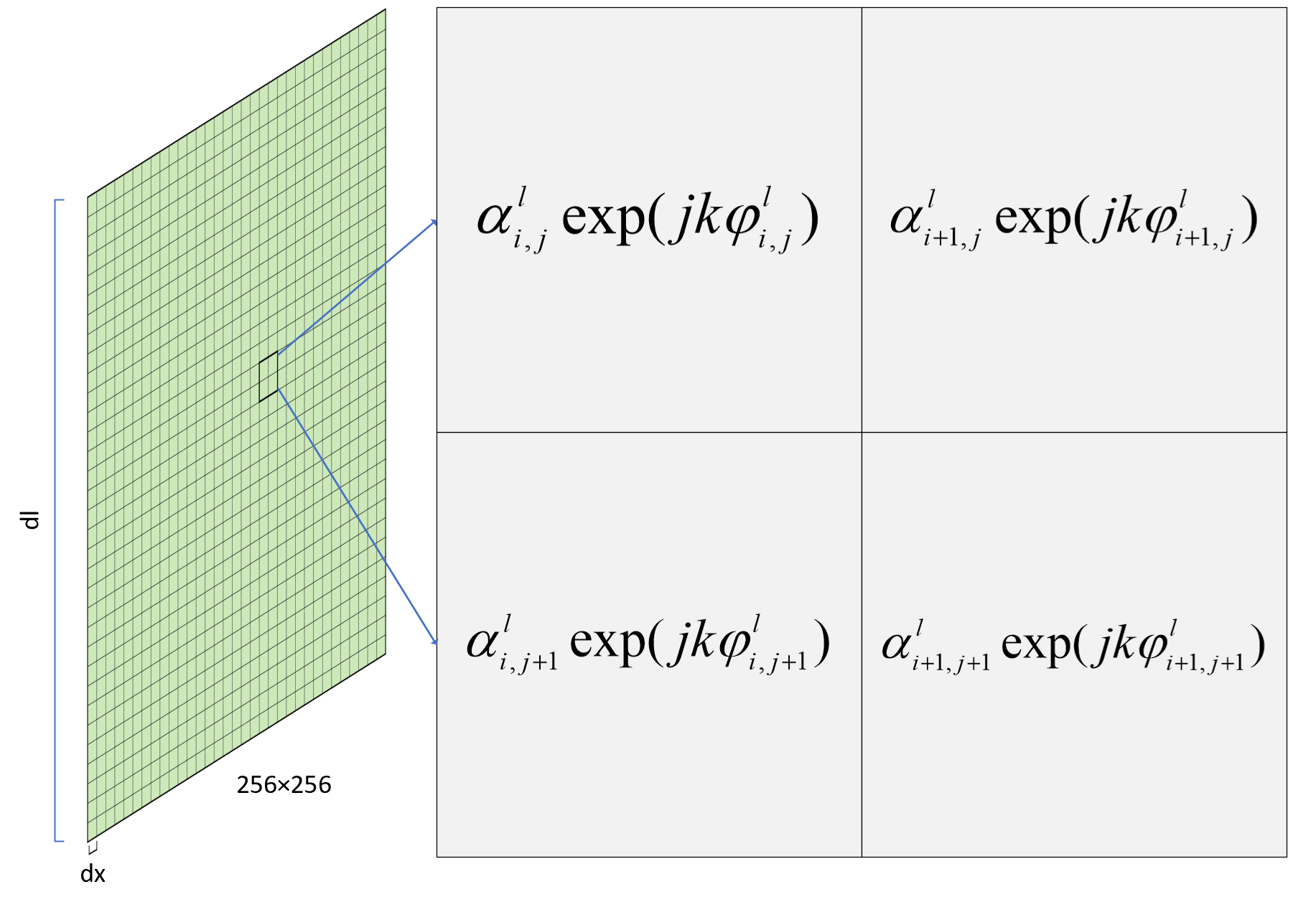}
\caption{The detail of modulation layer. 
 Each neuron in the modulation layer has a set of learnable parameters: amplitude and phase.}
\label{fig4}
\end{figure}

\subsection{Modulation Layer}

The modulation layer replaces the traditional convolutional layers found in conventional neural networks in EMWaveNet. While convolutional kernels typically consist of multiple channels, the modulation layer operates with a single channel. Instead of performing convolution operations, the modulation layer adjusts the amplitude and phase of the electromagnetic wave field. 

\begin{equation}
t(x,y,z)=a(x,y,z) e^{jk\phi(x,y,z)}
\label{eqn:MWaveNet}
\end{equation}

The modulation layer is depicted as illustrated in \Cref{fig4}. Each neuron within the modulation layer comprises a set of learnable parameters designed to modulate the amplitude and phase of the electromagnetic waves.

\begin{figure}[bp]
\centering
\includegraphics[width=3.7in]{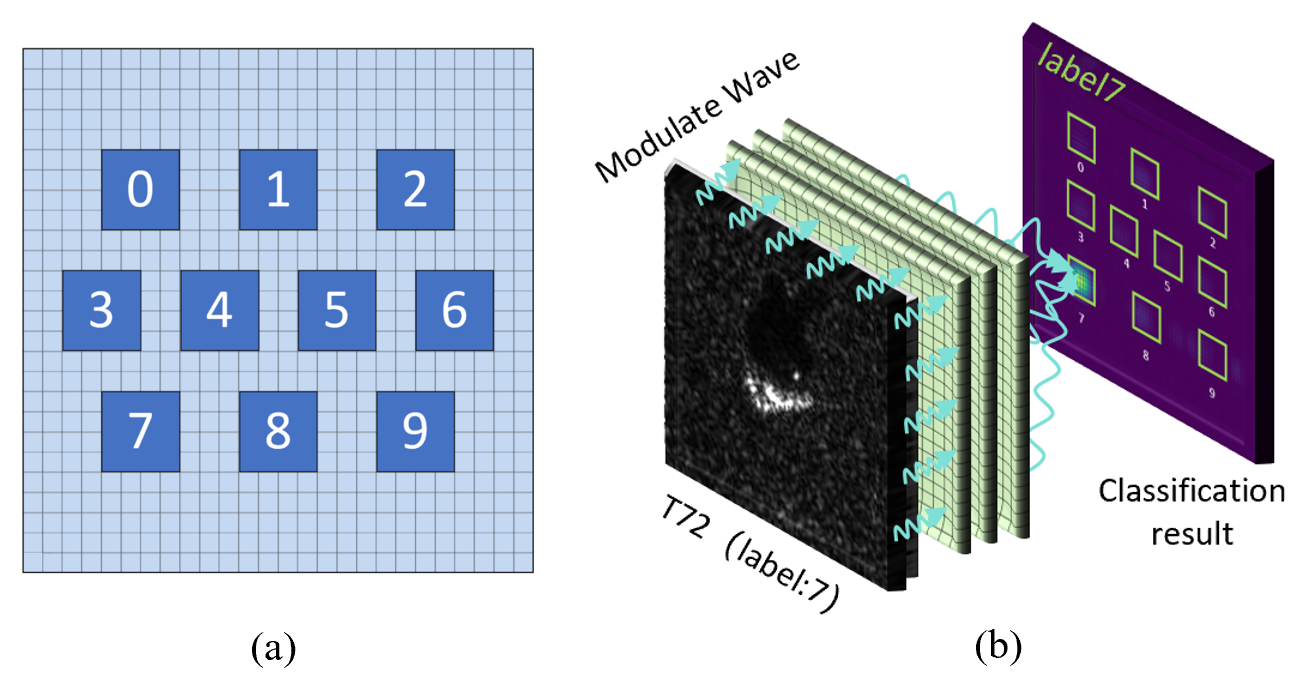}
\caption{The classification layer of EMWaveNet. (a) The Visualization
of classification of the classification layer. Taking the MSTAR dataset as an example, ten regions corresponding to ten target categories are selected on the classification layer. The classification task is realized based on the energy intensity of the microwave signal in the comparison area. (b) The visualization of the classification process on the MSTAR dataset for the T72 target category (label 7) using complex-valued SAR inputs.
}
\label{fig5}
\end{figure}

\subsection{Layer for Classification}

The output layer is configured as a classification layer to perform classification tasks. With a total of  \(M\) layers in the network (not including the input and output layers), the output layer is thus the \((M+1)\) layer. The microwave signal received at a neuron on the output layer is represented by \(m_i^{M+1}\), and the energy intensity \(s_i^{M+1}\) is given by

\begin{equation}
s_i^{M+1}=|m_i^{M+1}|^2.
\label{eqn:MWaveNet}
\end{equation}

As shown in \Cref{fig5} (a), regions are selected on the output layer according to categories \(c\). The input SAR image is classified based on the magnitude of the energy within the regions of the classification layer, as represented in \Cref{fig5}.

\subsection{Loss Function}

A novel loss function is developed named signal to noise ratio loss  \(L_{SNR}\)  to replace the conventional cross-entropy loss function, tailored for the classification layer of EMWaveNet. The energy for region consistent with the categories \(c\) is defined as \(s_{c}^{M+1}\), the other energy sum for the layer is represented as ~\(s^{M+1}-s_{c}^{M+1}\). The loss function is designed as follows

\begin{equation}
L_{SNR}=\sum_c\frac{(s^{M+1}-s_{c}^{M+1})}{s_c^{M+1}}
\label{loss}
\end{equation}

Throughout the training process, the value of the loss function continually decreases. This results in an increase in the signal energy value for the corresponding category region, while the values for other regions decrease.

\sethlcolor{yellow}
\subsection{Back Propagation}

The EMWaveNet is optimized by applying the back-propagation algorithm with the Adam optimization method using the loss function. Calculating the gradient of the loss \(L_{SNR}\) for both the phase \(\phi_i^l\) and amplitude \(a_i^l\) of each neuron in layer \(l\). These gradients are then used to iteratively update the phase and amplitude values across the network layers during each training epoch, with the Adam optimizer providing adaptive learning rates for more efficient convergence.

The gradient of \(L_{SNR}\) with respect to \(\phi_i^l\) can be written as

\begin{equation}
\frac{\partial L_{SNR}}{\partial \phi_i^l} = \sum_{i} \frac{\partial}{\partial \phi_i^l} \left( \frac{s_{i}^{M+1} - s_{i}^{M+1} \circ g_{i}}{s_{i}^{M+1} \circ g_{i}} \right)
\label{back1}
\end{equation}
where \(g_{i}\) is a mask associated with the target region (its elements do not depend on \(\phi_i^l\), and therefore it is constant). Set {\( N_i =s_{i}^{M+1} \circ (1-g_{i})\)}, {\( D_i =s_{i}^{M+1} \circ g_{i}\)}. Then, \Cref{back1} can be written as

\begin{align}
\frac{\partial  L_{SNR} }{\partial\phi_i^l}&=\sum_{i}\frac{D_{i}\frac{\partial N_{i}}{\partial\phi_{i}^{l}}-N_{i}\frac{\partial D_{i}}{\partial\phi_{i}^{l}}}{D_{i}^{2}} \nonumber \\
&=\sum_{i}\frac{D_i\left((1-g_i)\circ\frac{\partial s_i^{M+1}}{\partial\phi_i^l}\right)-N_i\left(g_i\circ\frac{\partial s_i^{M+1}}{\partial\phi_i^l}\right)}{D_i^2}
\label{back12}
\end{align}

To clearly represent the relationships between layers, we further define \(i_l\) for \(M+1 \geq l \geq M-L\), where \(1 \leq L < M\).
Set \(m_{i_{M+1}}^{M+1}\) as the weighted sum from all previous layers at the (\(i_{M+1}\))-th neuron in the output layer, the derivative of \(s_{i_{M+1}}^{M+1}\) with respect to \(\phi_i^l\) can be expressed as

\begin{equation}
\displaystyle\frac{\partial s_{i_{M+1}}^{M+1}}{\partial\phi_i^l}=2\operatorname{Re}\left((m_{i_{M+1}}^{M+1})^*\cdot\frac{\partial m_{i_{M+1}}^{M+1}}{\partial\phi_i^l}\right)
\label{back2}
\end{equation}
where \(\frac{\partial m_{i_{M+1}}^{M+1}}{\partial\phi_i^l}\) quantifies the sensitivity of the output field with respect to the phase values of neurons in the previous layers. 
\begin{figure*}[bp]
\centering
\includegraphics[width=7.2in]{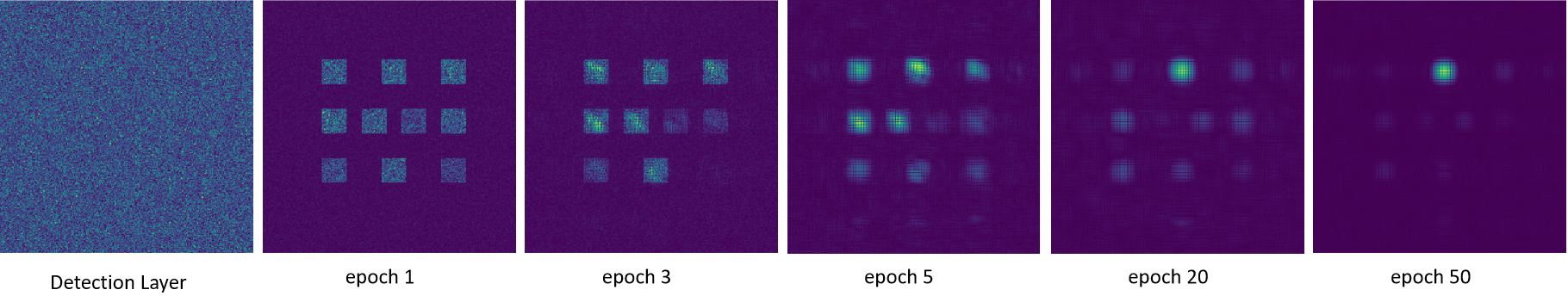}
\caption{The Visualization of Classification Detector Layer Fitting Process. As the number of learning epochs increases, the energy of the microwave signal gradually concentrates in the area to which the input SAR target belongs.}
\label{fig6}
\end{figure*}

 By applying the chain rule, The \(\frac{\partial m_{i_{M-L+1}}^{M-L+1}}{\partial \phi_{i_{M-L}}^{M-L}}\) can be written:

\begin{equation}
\frac{\partial m_{i_{M-L+1}}^{M-L+1}}{\partial \phi_{i_{M-L}}^{M-L}}= \frac{\partial m_{i_{M-L+1}}^{M-L+1}}{\partial n_{i_{M-L}}^{M-L}}  \cdot\frac{\partial n_{i_{M-L}}^{M-L}}{\partial t_{i_{M-L}}^{M-L}}\cdot\frac{\partial t_{i_{M-L}}^{M-L}}{\partial \phi_{i_{M-L}}^{M-L}}
\label{back44}
\end{equation}

The gradient can be recursively calculated for each layer, as shown in the following equation:

\begin{align}
\frac{\partial m_{i_{M+1}}^{M+1}}{\partial \phi_{i}^{l}}&=\frac{\partial m_{i}^{M+1}}{\partial \phi_{i_M}^{M}}\cdot\frac{\partial m_{i_M}^{M}}{\partial m_{i_{M-1}}^{M-1}}\cdots\frac{\partial m_{i_{M-L+1}}^{M-L+1}}{\partial \phi_{i_{M-L}}^{M-L}}  \\[7pt]
&= \sum_{i_{M}} h_{i_{M},i_{M+1}}^{M}\cdot t_{i_M}^M\cdot\sum_{i_{M-1}} h_{i_{M-1},i_{M}}^{M-1}\cdot t_{i_{M-1}}^{M-1} \nonumber \\[7pt]
&\quad \cdots j \cdot h_{i_{M-L},i_{M-L+1}}^{M-L}\cdot t_{i_{M-L}}^{M-L} \cdot m_{i_{M-L}}^{M-L}\nonumber
\label{back4}
\end{align}

Similarly, the gradient of the loss with respect to the amplitude \(a_i^l\) of a given layer \(l\) is computed as follows

\begin{equation}
\frac{\partial L_{SNR}}{\partial a_i^l}=\sum_k\frac2{g_{i_{M+1}}^{M+1}}\operatorname{Re}\left(m_{i_{M+1}}^{M+1}\cdot\frac{\partial m_{i_{M+1}}^{M+1}}{\partial a_i^l}\right)
\label{back5}
\end{equation}

\begin{equation}
\frac{\partial m_{i_{M-L+1}}^{M-L+1}}{\partial a_{i_{M-L}}^{M-L}}=\frac{\partial m_{i_{M-L+1}}^{M-L+1}}{\partial n_{i_{M-L}}^{M-L}}\cdot\frac{\partial n_{i_{M-L}}^{M-L}}{\partial t_{i_{M-L}}^{M-L}}\cdot\frac{\partial t_{i_{M-L}}^{M-L}}{\partial a_{i_{M-L}}^{M-L}}
\label{back66}
\end{equation}

\begin{align}
\frac{\partial m_{i_{M+1}}^{M+1}}{\partial a_{i}^{l}}&=\frac{\partial m_{i_{M+1}}^{M+1}}{\partial a_{i_M}^{M}}\cdot\frac{\partial m_{i_M}^{M}}{\partial m_{i_{M-1}}^{M-1}}\cdots\frac{\partial m_{i_{M-L+1}}^{M-L+1}}{\partial a_{i_{M-L}}^{M-L}}  \\[7pt]
&= \sum_{i_{M}} h_{i_{M},i_{M+1}}^{M}\cdot t_{i_M}^M\cdot\sum_{i_{M-1}} h_{i_{M-1},i_{M}}^{M-1}\cdot t_{i_{M-1}}^{M-1} \nonumber \\[7pt]
&\quad \cdots h_{i_{M-L},i_{M-L+1}}^{M-L}\cdot \exp(j\phi_{i_{M-L}}^{M-L}) \cdot m_{i_{M-L}}^{M-L} \nonumber
\label{back6}
\end{align}

\section{Experimental Results}
\subsection{Dataset and Experimental Settings}

\begin{table}[bp]
\caption{\textbf{MSTAR DATASET DETAILS}}
\centering
\begin{tabular}{cccccc}
\toprule
Categories&Series&Depression&Trainset&Testset&Total \\
\midrule
BMP2 & 9563 & 17° & 186 & 47 & 233 \\
BTR70 & c71 & 17° & 186 & 47 & 233 \\
T72 & 132 & 17° & 186 & 46 & 232 \\
T62 & A51 & 17° & 239 & 60 & 299 \\
BRDM2 & E71 & 17° & 238 & 60 & 298 \\
BTR60 & 7532 & 17° & 205 & 51 & 256 \\
ZSU23/4 & d08 & 17° & 240 & 59 & 299 \\
D7 & 13015 & 17° & 240 & 59 & 299 \\
ZIL131 & E12 & 17° & 240 & 59 & 299 \\
2S1 & B01 & 17° & 240 & 59 & 299 \\
\bottomrule
\label{t1}
\end{tabular}
\end{table}

In this section, experiments were conducted on the publicly available MSTAR dataset and the self-developed Qilu-1 dataset. Initially, target recognition tasks were performed on both datasets. Subsequently, the performance of EMWaveNet and other traditional networks was evaluated under conditions of overlap, overlay, masking, and interference. Finally, an end-to-end complex-valued synthetic aperture radar
automatic target recognition (SAR-ATR) algorithm was developed, enabling automatic detection and recognition tasks in complex interference scenarios.

\subsubsection{MSTAR}
The MSTAR dataset collected by Sandia National Laboratories SAR sensor platform is selected as the baseline for the study [37]. This SAR sensor operates in the X-band and uses HH polarization and the data set is widely used in SAR automatic target recognition research. Each data point is a complex-valued image that can be decomposed into amplitude and phase. The image has a resolution of 0.3 meters × 0.3 meters and a size of 128 × 128 pixels, covering 360° in all directions. The MSTAR dataset contains ten different military vehicle types, namely rocket launcher: 2S1, armored personnel carrier: BMP2, BRDM2, BTR70, BTR60, bulldozer: D7, tank: T62, T72, truck: ZIL131 and air defense unit: ZSSU234.

In this experiment, images with the depression angle of 17° were selected as the sample dataset, comprising a total of 2747 images. Of these, 2200 images were utilized for training purposes, while 547 images were designated for testing. For specific details, refer to \Cref{t1}.

\begin{table}[bp]
\caption{\textbf{Qilu-1
DATASET DETAILS}}
\centering
\begin{tabular}{cccc}
\toprule
Categories&Trainset&Testset&Total \\
\midrule
A & 92 & 4 & 96  \\
B & 47 & 4 & 51  \\
C & 57 & 4 & 61  \\
\bottomrule
\label{t2}
\end{tabular}
\end{table}

\subsubsection{Qilu-1}
Additionally, a SAR aircraft slice dataset was independently developed to corroborate the findings of this research. Data for this dataset were derived from the Qilu-~1 satellite, a high-resolution remote sensing satellite autonomously developed by China, equipped with cutting-edge remote sensing imaging technology. It offers high-resolution and high-precision Earth observation data, encompassing an extensive collection of remote sensing images and geographic information data, with broad geographical coverage. A dataset comprising complex-valued SAR image slices of aircraft was constructed using 11 scenes from airports. 

In this experiment, 196 images Experiments were utilized for training purposes, while 12 images were designated for testing. The detailed information is referred to the \Cref{t2}.

\begin{figure*}[!h]
\centering
\includegraphics[width=6in]{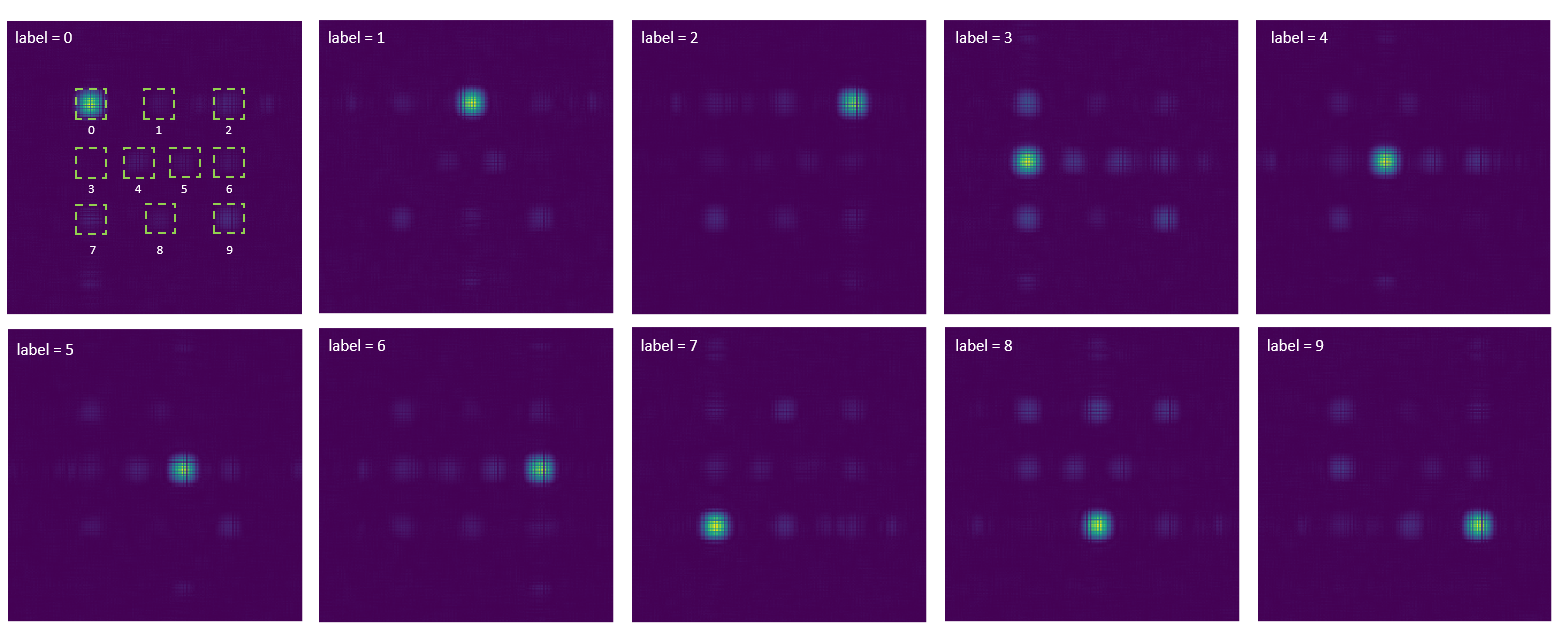}
\caption{Visualization results of the classification detection layer for target recognition of different categories in the MSTAR dataset.}
\label{fig7}
\end{figure*}

\begin{table}[bp]
\caption{\textbf{EMwaveNet Hyperparameter Settings for Different Datasets}}
\centering
\begin{tabular}{cccccccc}
\toprule
\(\Theta\) & \(f\) & \(\lambda\) & \(M\) & \(N\) & \(d\) & \(dx\) & \(dl\) \\
\midrule
MSTAR & 9.6\text{GHz} & 0.03\text{m} & 5 &256& 0.3\text{m} & 0.0001\text{m}&0.256\text{m} \\
Qilu-1 & 16.7\text{GHz} & 0.02\text{m} & 5 &256& 0.3\text{m} & 0.0001\text{m}&0.256\text{m}  \\
\bottomrule
\label{t3}
\end{tabular}
\end{table}

\subsubsection{Experimental Settings}
All input images for the experiments were resized to 256\(×\)256 pixels. The training was optimized using the Adaptive Moment Estimation (Adam) optimizer with an initial learning rate of 0.1 and conducted over a total of 200 epochs. Additionally, our network employs a learning rate decay strategy, wherein the learning rate is periodically adjusted using the StepLR scheduler in PyTorch. Specifically, the learning rate is reduced by a factor of 0.5 every 20 training steps, thereby enhancing the model's performance and stability. All experiments are conducted on a workstation of 64-bit Linux operating system with NVIDIA RTX 3090.

\subsection{Target Recognition}

\subsubsection{Hyperparameter Calculation}

The hyperparameters required for the initialization of the EMWaveNet network can be calculated based on the radar parameters provided by the dataset. A set of hyperparameters \(\Theta=[f,\lambda,M,N,d,dx,dl]\) needs to be determined. \(f\) is the radar operating frequency, \(\lambda\) is the wavelength, \(M\) represents the number of modulation layers, and \(N \times N\) is the resolution of the image. \(d\) denotes the distance between layers, \(dx\) is the spatial sampling interval, which must be less than half the wavelength according to the Nyquist theorem. \(dl\) is the length of the layer, with \(dl = N \times dx\). The set of hyperparameters used in the experiments of this article is shown in \Cref{t3}.

\begin{table}[!t]
\caption{\textbf{Classification Performance Comparison of Various Models on MSTAR}}
\centering
\begin{threeparttable}
\begin{tabular}{cccc}
\toprule
&Accuracy(\%) &Parameters(M)&Flops(G)\\
\midrule
AlexNet&99.82&61.20&3.16\\
Vgg19&99.64&143.7&19.61\\
ResNet18&99.37&11.69&2.38\\
ResNet50&99.10&25.56&8.01\\
ResNext50& 99.46&25.04&8.23 \\
ResNext101& 98.19&44.55&14.68 \\
DenseNet121& 98.92&25.60&4.12 \\
Vision-Transformer& 99.64& 86.61& 17.60\\
Swin-Transformer& 99.82& 88.00& 15.40\\
\midrule
CV-AlexNet & 99.64 & 122.4 & 6.32 \\
CV-Vgg19 & \textbf{99.82} & 287.34 & 39.22\\
CV-ResNet18 & 99.58 & 23.38 & 19.43 \\
CV-ResNeXt50& 99.64& 51.12 & 16.02\\
CV-DenseNet121& 99.28& 50.08 & 16.46\\
Ours & 99.58 & \textbf{1.31} & \textbf{0.12} \\

\bottomrule
\label{t4}
\end{tabular}
\begin{tablenotes}
\footnotesize
\item[1] In this article, ``CV" stands for complex value.
\end{tablenotes}            
\end{threeparttable}
\end{table}

\subsubsection{Recognition Results}
Experiments were conducted on the MSTAR and Qilu-1 datasets to validate the recognition performance of EMWaveNet. \Cref{fig6} visualizes the fitting process of the detector, demonstrating that throughout the training phase, with the reduction of the loss function, the signal energy within the detection layer becomes focused in the area corresponding to the respective category of the target. \Cref{fig7} displays the visualization of classification results for ten target types from the MSTAR dataset on the detector, indicating that inputs from different categories prompt activation in varying regions of the detection layer. As shown in the \Cref{t4}, several neural networks that have shown exceptional performance on classification tasks were selected for comparison with our algorithm. Our model achieves classification accuracy comparable to traditional networks with fewer parameters and reduced computational complexity, requiring fewer floating-point operations (flops) per inference. Notably, the entire classification process and parameters of our model have specific physical significance.

\begin{table}[bp]
\caption{\textbf{Accuracy of Models with Different Layer Depth}}
\centering
\begin{tabular}{cccccc}
\toprule
 No. of Layers &1&3&5&10&15 \\
\midrule

Acc &45.17& 98.91 & 99.34 & \textbf{99.58}& 99.16\\
\bottomrule
\label{t5}
\end{tabular}
\end{table}

\begin{table}[!t]
\caption{\textbf{Accuracy of Models with Different Modulation Layer Size}}
\centering
\begin{tabular}{cccccccc}
\toprule
 Mod. Size &64&88&128&192&256&320&512 \\
\midrule

Acc &98.01& 98.73 & 98.92 & 99.28&\textbf{99.58}& 99.10&99.46\\
\bottomrule
\label{t52}
\end{tabular}
\end{table}

\begin{table}[bp]
\caption{\textbf{Accuracy of Models with Different Combinations of Learnable Parameters}}
\centering
\begin{tabular}{ccc}
\toprule
Learnable Parameters&Phi&Amp \& Phi \\
\midrule

Acc & 98.19 & \textbf{99.58} \\
\bottomrule
\label{t6}
\end{tabular}
\end{table}

\subsubsection{Model Configuration}

\paragraph{The Number of Modulation Layers M}

Similar to other deep learning models, the number of modulation layers directly impacts the model's expressive capacity. As the number of layers increases, the network can learn more complex features and patterns. Optimizing the depth of EMWaveNet requires careful consideration of the trade-off between model complexity, computational efficiency, and the ability to capture and utilize the intricate details of microwave signal interactions.
\begin{figure}[!t]
\centering
\includegraphics[width=3.4in]{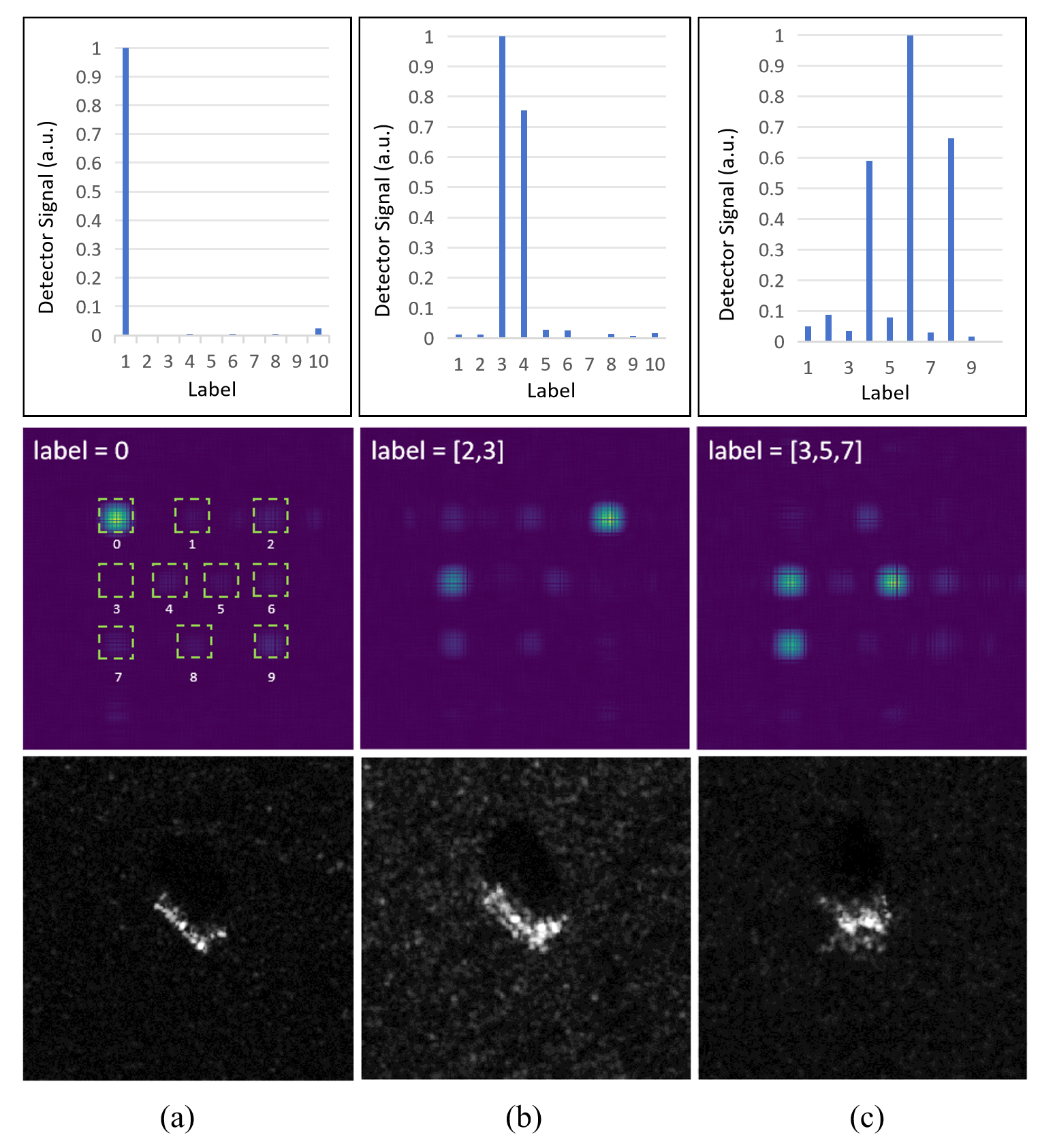}
\caption{Visualization of Recognition results from the coherent superposition of different complex-valued SAR targets. The first row shows the normalized values of signal energy in the 10 regions of the detector. The Second row displays the visualization of the classification detection layer, while the third row shows the amplitude graph of the targets' coherent superposition. (a) represents a single target, labeled as 0. (b) depicts the overlapping of two targets, categories 2 and 3. (c) illustrates the blending of three types of targets, categories 3, 5, and 7.}
\label{fig888}
\end{figure}

\begin{figure*}[bp]
\centering
\includegraphics[width=6.5in]{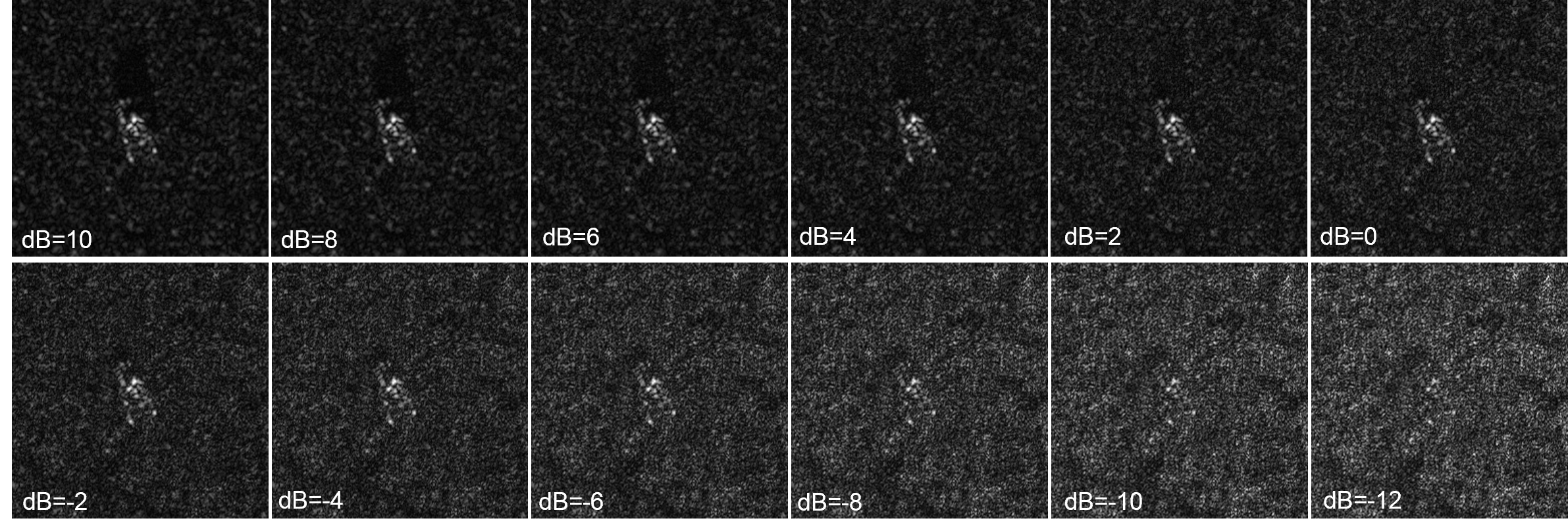}
\caption{SAR target images under forest interference with different SNR.}
\label{fig9}
\end{figure*}
As indicated in \Cref{t5}, an increase in the number of layers is associated with improved accuracy. However, beyond 15 layers, a decrease in accuracy is observed. This trend is attributed to the increasing complexity of network, which leads to overfitting. In the remaining experiments of the article, a depth of 5 layers was uniformly selected for the network.
\begin{table}[!t]
\caption{\textbf{Overlapped Target Recognition Results of Coherent Superposition of Different Numbers of Targets}}
\centering
\begin{tabular}{cccc}
\toprule
    Overlap Target Count&1&2&3\\
\midrule
AlexNet& 100&15.56&1.67\\
Vgg19& 100&2.22&0.83\\
ResNet18& 100&2.22&0.83\\
ResNext50& 100&2.22&0.83\\
DenseNet121& 100&2.22&0.83\\
Swin-Transformer& 100&22.22&3.33\\
CV-AlexNet & 100 & 31.15  & 7.50 \\
CV-Vgg19 & 100 & 2.22 &0.83 \\
CV-ResNet18 & 100 & 2.22 &0.83 \\
CV-ResNeXt50& 100&2.22&0.83\\
CV-DenseNet121& 100&2.22&0.83\\
Ours & 100 & \textbf{100} & \textbf{78.33} \\
\bottomrule
\label{t7}
\end{tabular}
\end{table}

\paragraph{Modulation Layers Size}

An increase in the modulation layer size results in more parameters and higher computational cost per layer, thereby enhancing the network's capacity and learning ability. 

As shown in \Cref{t52}, experiments were conducted with different sizes of the modulation layer. it can be observed that as the modulation layer size increases from 64 to 256, the accuracy gradually improves, with a significant increase from 98.01\% to 99.58\%. This indicates that increasing the modulation layer size helps the model better learn the features in the data, thereby improving classification accuracy. When the modulation layer size is 256, the model achieves the best accuracy with relatively efficient computational cost. Increasing the modulation layer size further (such as 320 or 512) may lead to increased computational complexity, with limited performance gains, and could even introduce issues such as overfitting. Therefore, in our experiments, choosing a modulation layer size of 256 offers a balance between accuracy and computational efficiency.                                                

\paragraph{Learnable Parameters}

In the modulation layers of EMWaveNet, there are two learnable parameters:  \{\(a_i^l\), \(\phi_i^l\)\}, which modulate the amplitude and phase of the electromagnetic waves, respectively. An experiment was conducted to evaluate the impact of these coefficients on network performance. The results clearly demonstrate that simultaneously learning Amp and Phi significantly improves network effectiveness compared to learning each parameter separately, as shown in  \Cref{t6}.

\begin{figure*}[bp]
\centering
\includegraphics[width=5.5in]{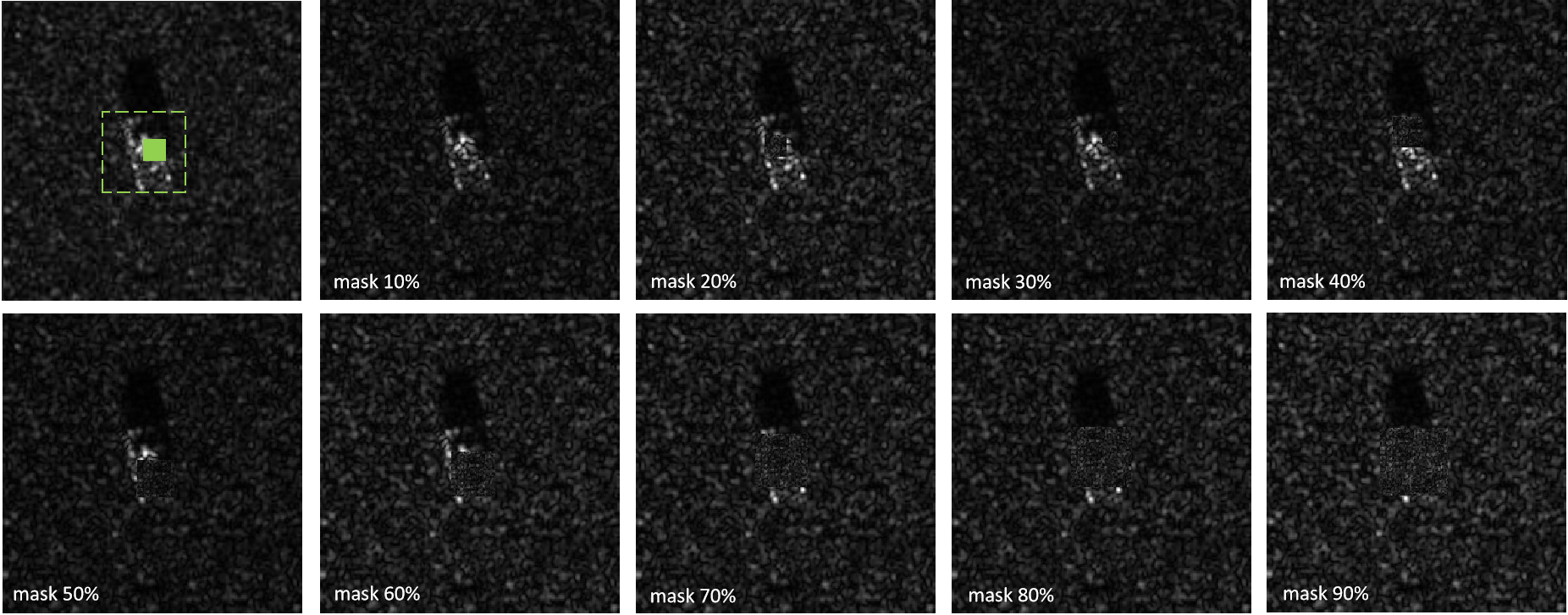}
\caption{SAR images with random-sized forest interference at a 0dB SNR.}
\label{fig11}
\end{figure*}

\begin{figure*}[!t]
\centering
\includegraphics[width=7.2in]{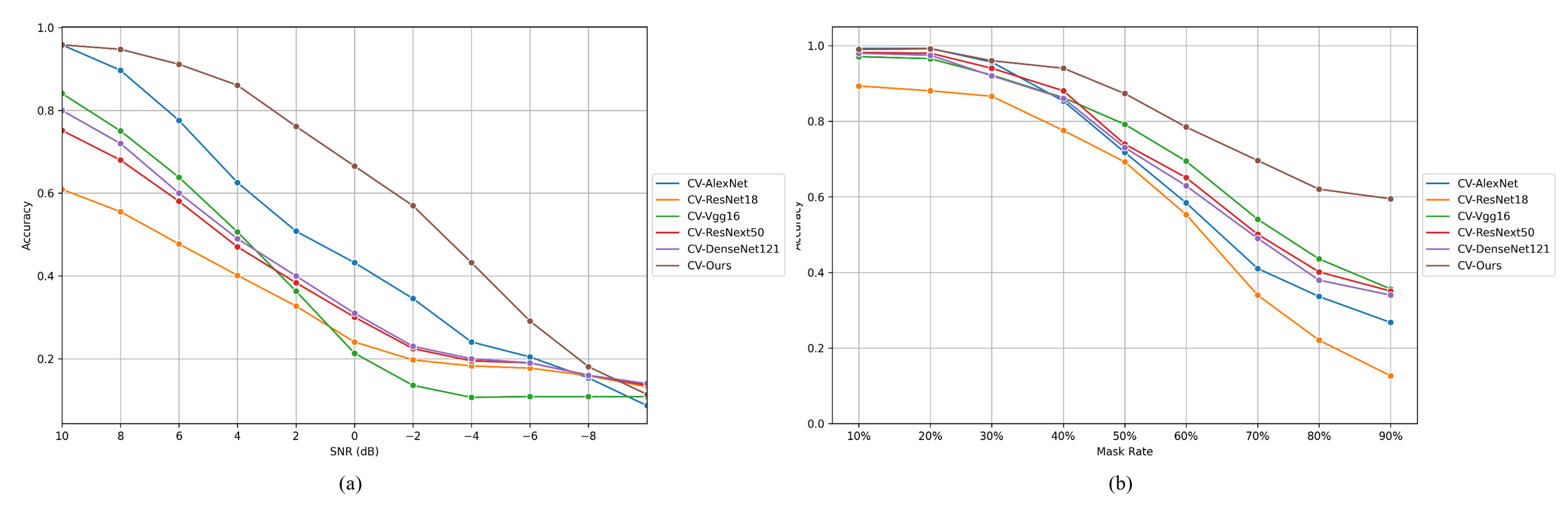}
\caption{Model accuracy across different forest interference. (a) The recognition results for forest interference at different SNR. (b) The recognition results for random-sized interference at a 0dB SNR.}
\label{fig10}
\end{figure*} 

\begin{figure}[!t]
\centering
\includegraphics[width=3.1in]{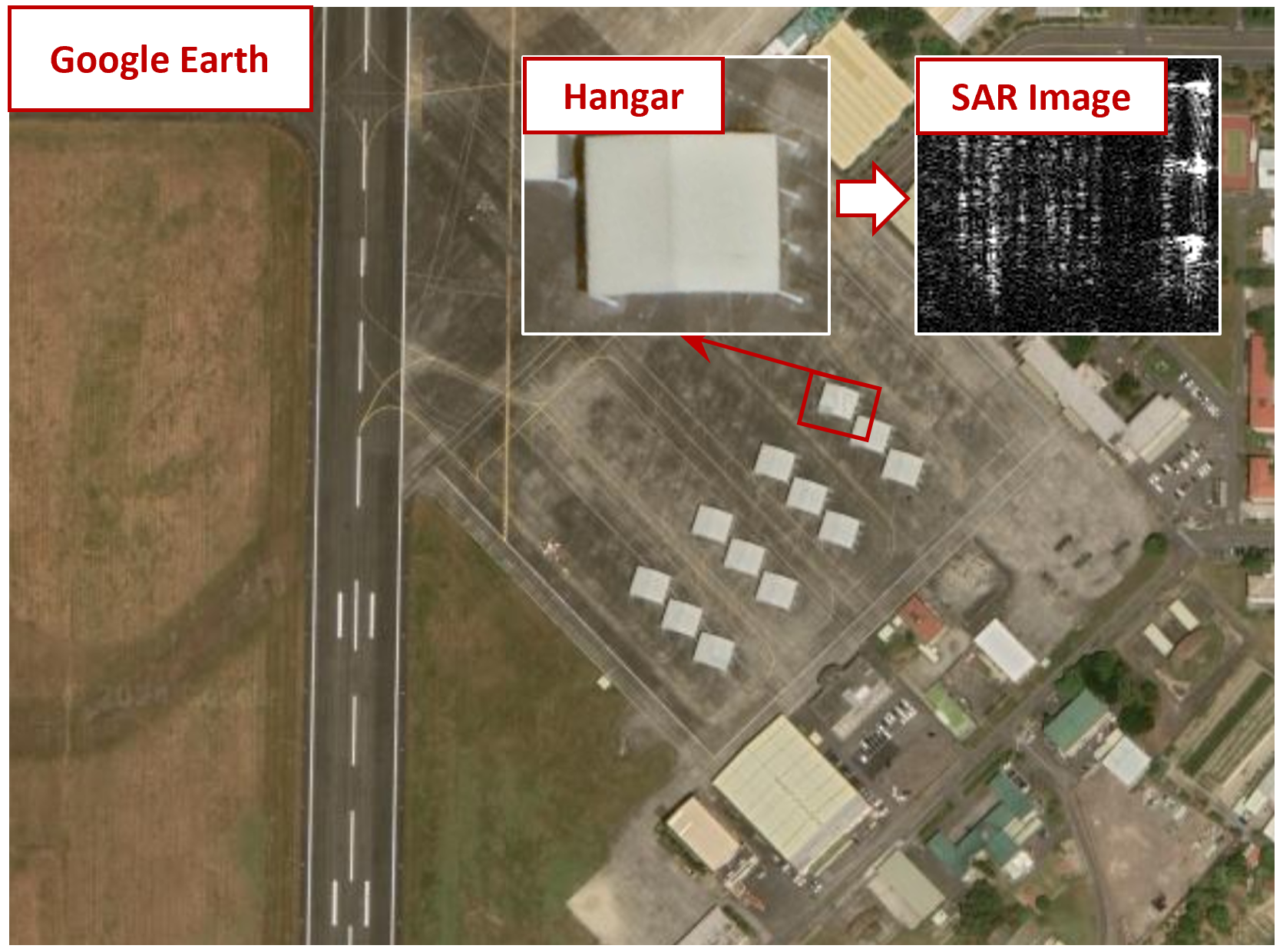}
\caption{Optical image of an airfield with some hangars. }
\label{fig1313}
\end{figure}

\begin{figure*}[!t]
\centering
\includegraphics[width=5.7in]{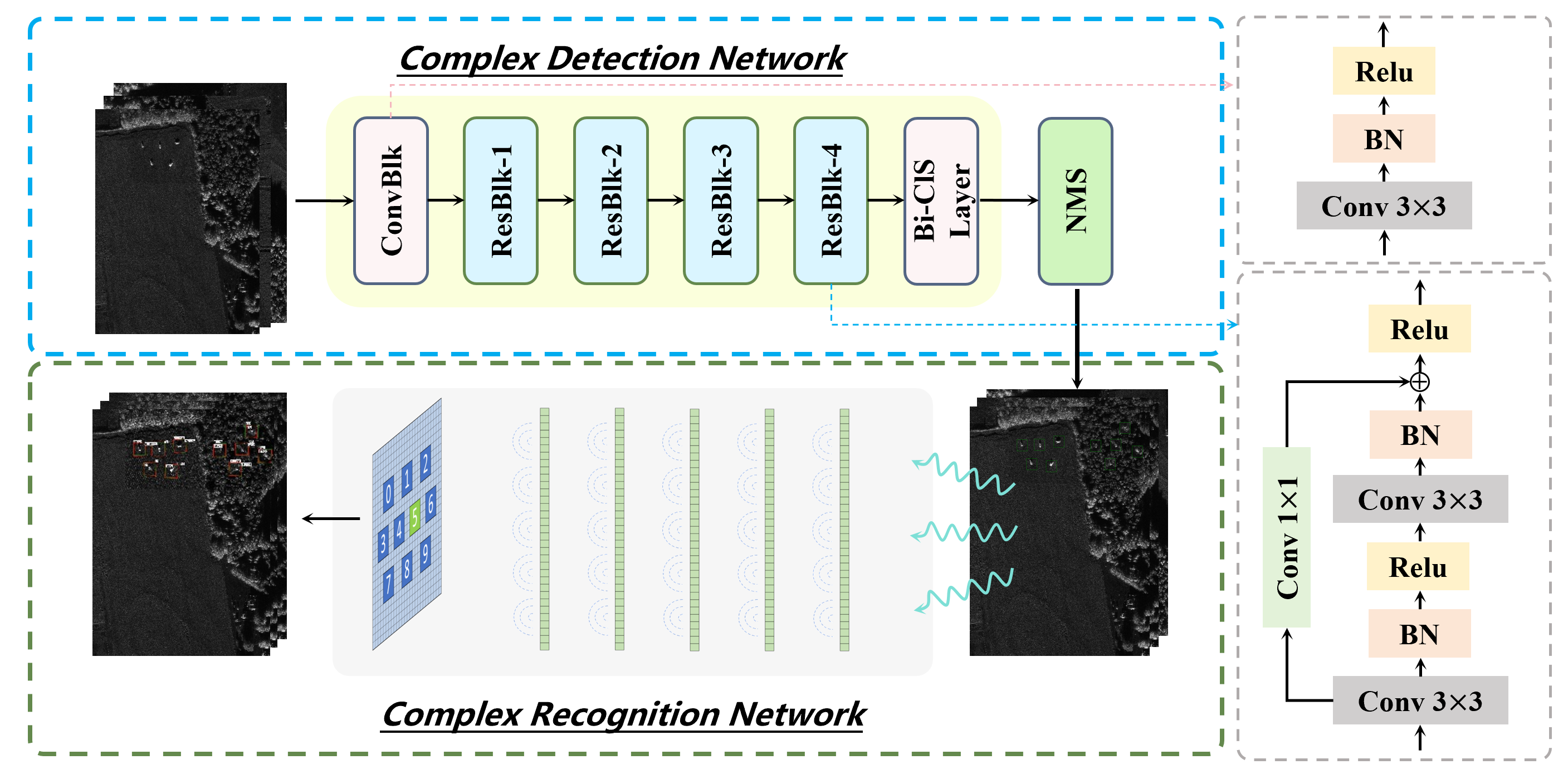}
\caption{Overall Framework of the End-to-End Complex-Valued SAR-ATR System. This system is composed of two main parts: a detection network and a recognition network. The input is complex-valued SAR scene images, and the output is the recognition results of the targets.}
\label{fig1311}
\end{figure*}

\begin{figure*}[!t]
\centering
\includegraphics[width=7.2in]{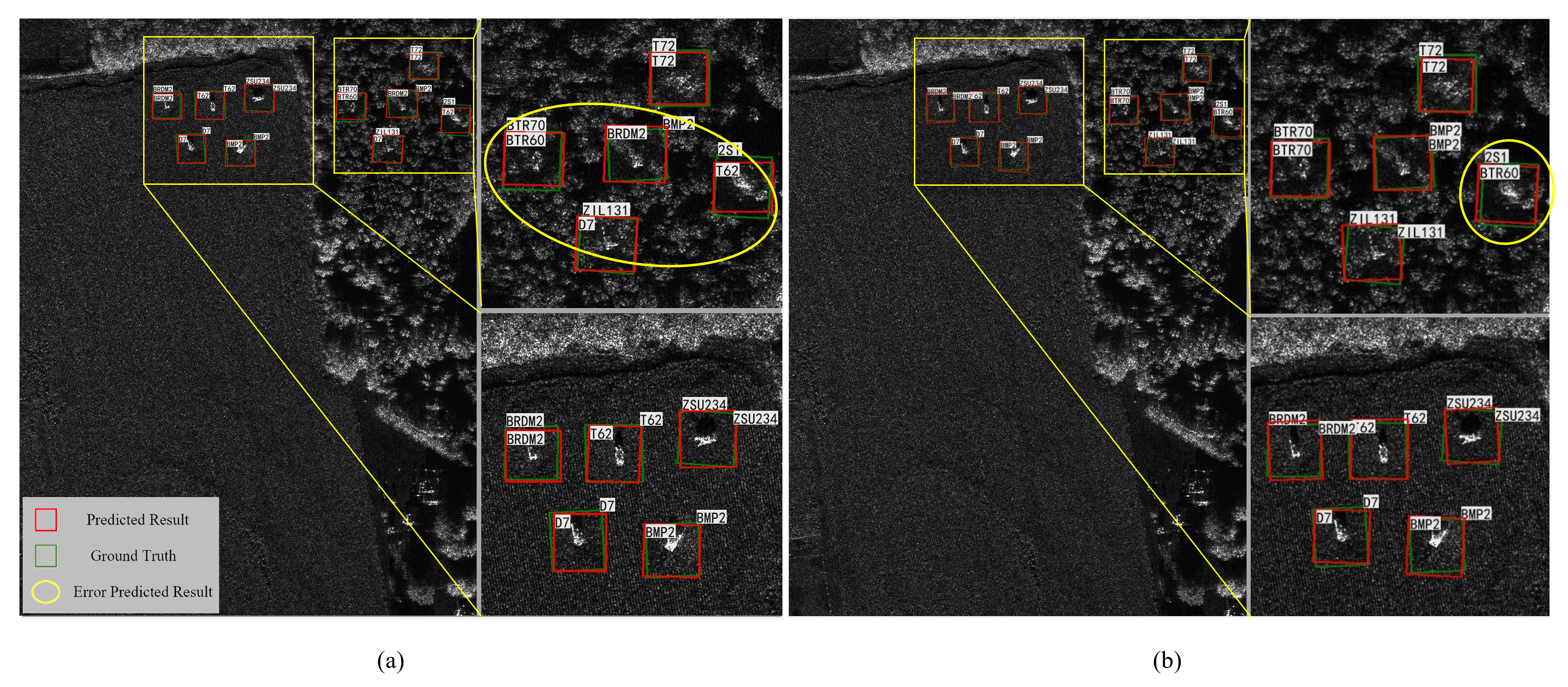}
\caption{Detection and Recognition Results in Complex Interference SAR Scenarios. (a) The backbone of the recognition module is CV-ResNet18. (b) The backbone of the recognition module is EMWaveNet. In the images, green boxes represent the ground truth, red boxes indicate the detection and recognition results, and yellow circles highlight the erroneous recognitions.}
\label{fig1312}
\end{figure*}

\subsection{Interference Experiment}
\subsubsection{Target Coherent Superposition}
\begin{table*}[!t]
\caption{\textbf{Comparison of Classification Performance of Various Models on MSTAR\\under Different SNR Levels with Forest Noise Overlay}}
\centering
\begin{tabular}{cccccccccccc}
\toprule
SNR&-10dB&-8dB&-6dB&-4dB&-2dB&-0dB&2dB&4dB&6dB&8dB&10dB\\
\midrule
CV-AlexNet & 8.68 & 15.37 & 20.43 &24.05& 34.54 & 43.22 & 50.81 & 62.57 & 77.58 & 89.69 & 95.67 \\
CV-Vgg19 & 10.84 & 10.84 & 10.84 & 10.67 & 13.56 & 21.34 & 36.35 & 50.63 & 63.83 & 75.05 & 84.09 \\
CV-ResNet18 & \textbf{13.20} & 15.91 & 17.72 & 18.26 & 19.71 & 24.05 & 32.73 & 40.14 & 47.74 & 55.52 & 60.94 \\
CV-ResNext50&13.56&16.04&18.99&19.48&22.44&30.05&38.37&47.04&58.04&67.99&75.17	\\
CV-DenseNet121&14.00&16.56&19.00&20.09&22.99&31.01&40.04&49.15&60.06&72.19&80.22
\\

Ours & 11.47 & \textbf{18.14} & \textbf{29.13} & \textbf{43.21} & \textbf{60.05} & \textbf{66.56} & \textbf{76.17} & \textbf{86.02} & \textbf{91.17} & \textbf{94.74} & \textbf{95.84}\\
\bottomrule
\end{tabular}
\end{table*}

\begin{table*}[!t]
\caption{\textbf{Comparison of Classification Performance of Various Models on MSTAR\\under Different Ratio with 0dB Forest Noise masking}}
\centering
\begin{tabular}{cccccccccc}
\toprule
Masking Ratio&10\%&20\%&30\%&40\%&50\%&60\%&70\%&80\%&90\%\\
\midrule
CV-AlexNet & \textbf{99.28} & \textbf{99.28} & 95.66 &85.35& 71.79 & 58.41 & 41.05 & 33.63 & 26.76\\
CV-Vgg19 & 97.11 & 96.56 & 92.22 & 86.26 & 79.20 & 69.44 & 54.07 & 43.58 & 35.62 \\
CV-ResNet18 & 89.33 & 88.17 & 86.62 & 77.58 & 69.26 & 55.33 & 33.99 & 22.06 & 12.66  \\
CV-ResNext50&98.20&98.01&94.03&88.07&73.96&65.10&50.10&40.14&35.08\\
CV-DenseNet121&98.01&97.47&92.04&86.08&73.06&62.93&49.01&37.97&34.00\\
Ours & 99.01 & 99.19 & \textbf{96.02} & \textbf{94.03} & \textbf{87.34} & \textbf{78.48} & \textbf{69.62} & \textbf{62.03} & \textbf{59.49} \\
\bottomrule
\end{tabular}
\end{table*}

Samples from ten categories were randomly
selected from the MSTAR trainset and coherently superimposed to create overlapping target images. As illustrated in \Cref{fig888}, the second row of images represents overlapping targets, with "label" denoting their respective categories. It can be observed that the features of the targets are blended together, making them difficult to recognize.  The first row presents a visualization of the energy intensity of the classification detection layer of the network. It is evident that the EMWaveNet can easily identify the categories of the constituent targets from the overlapped images. As shown in the \Cref{t7}, traditional deep learning networks can almost not handle overlapped target images. In contrast, EMWaveNet accurately recognizes overlapping target categories, demonstrating its superior de-overlapping capabilities.

\subsubsection{Forest Interference}

In this study, forest SAR images that align with the spectral band and polarization mode of the MSTAR dataset were employed to introduce interference. These images were cropped to match the dimensions of the original targets and then coherently superimposed onto the original complex-valued images, creating interference and simulating the environmental conditions of targets obscured by forest canopy. To assess the impact of forest noise on recognition outcomes, varying levels of forest interference were simulated using different signal-to-noise ratios (SNRs), defined as follows:

\begin{equation}
SNR=10\log_{10}\frac{P_{\text{singal}}}{P_{\text{noise}}}=10\log_{10}\frac{\sum x^2}{\sum n^2}.
\label{e26}
\end{equation}

\Cref{fig9} illustrates the variation in target images under different SNR conditions of forest interference. It can be observed that as the SNR increases, the proportion of noise escalates, leading to a more extensive coverage of the targets by noise. \Cref{fig10} (a) reveals the recognition outcomes under various SNR conditions of forest interference. The performance of EMWaveNet in identifying targets beneath forest canopy has shown a significant improvement compared to traditional neural networks. This further substantiates the enhanced robustness and accuracy of EMWaveNet when tasked with recognizing targets under conditions of noise interference. 

\subsubsection{Random Mask Forest Interference}

To simulat more realistic scenarios of targets under the forest, noise characterized by random dimensions is designed to interfere with the targets. The noise interference is constrained to an 88\(\times\)88 area in the center to ensure that all interference covers the target. As shown in \Cref{fig11}, forest noise of different sizes is coherently superposed with the target to simulate a more realistic forest semi-interference scene. \Cref{fig10} (b) shows the recognition results of different models in the semi-disturbed forest situation. EMWaveNet exhibits better recognition performance as the interference size increases than traditional convolutional neural networks. The greater the interference noise, the more robust the EMWaveNet is.
\subsubsection{Hangar Interference}

An experiment involving overlaid hangar interference was designed for the self-built Qilu dataset. As depicted in \Cref{fig1313}, this scenario illustrates the situation where aircraft parked within the hangars with their features disrupted and rendered unrecognizable due to interference. Experiments were conducted with the test set aircraft samples overlaid with hangar noise. As shown in \Cref{t9}, EMWaveNet is capable of recognizing more aircraft targets disturbed by hangar interference compared to other models. This demonstrates that the algorithm proposed in this study not only possesses high explainability throughout its process but also exhibits strong robustness.

\begin{table}[ht]
\caption{\textbf{The recognition results of the models on the Qilu-1 dataset}}
\centering
\begin{tabular}{ccccc}
\toprule
    Model&CV-AlexNet&CV-Vgg19&CV-ResNet18&Ours\\
\midrule
No Hangar  & 100 & 100 & 100  & 100\\
Hangar & 41.6 & 33.3 & 33.3 & \textbf{83.3} \\
\bottomrule
\label{t9}
\end{tabular}
\end{table}

\subsection{SAR-ATR in Complex Interference Scene}

The public MSTAR dataset includes not only target chips but also scene images without targets. Both the scene images and target chips are acquired by the same SAR sensor, with a resolution of 0.3m and HH polarization. Therefore, superimposing the target chips onto the scenes is justifiable. SAR scene data has been simulated with embedded targets in complex interference scenarios for the experiments.

A complex-valued end-to-end SAR-ATR algorithm was designed for the task of target detection and recognition in complex SAR scenes. The algorithm architecture is illustrated in \Cref{fig1311}. The input to the network is complex-valued SAR scene images, and the output is the recognition results. The network consists of two main components: a complex-valued detection network and a complex-valued recognition network. The backbone of detection network is a binary classification network based on CV-ResNet18, which performs target and background classification through sliding window traversal of the image. Subsequently, the results of the binary classification are fed into the Non-Maximum Suppression (NMS) network, which is tasked with selecting the optimal bounding boxes. The recognition network is a complex-valued EMWaveNet, which takes the detection results and processes them to produce the final recognition outcomes.

Detection and Recognition Results are illustrated in \Cref{fig1312}. The results demonstrate that for targets not obscured by forest cover, both networks accurately identify the target categories. However, for targets obscured by forest noise, the CV-ResNet18 correctly identifies only one target, whereas the EMWaveNet correctly identifies four targets. This underscores the strong noise resistance and robustness of the proposed network in complex interference SAR scenarios.

\subsection{Disscuss}

In the experiments conducted in this Section, EMWaveNet demonstrated superior deblurring performance, capable of accurately identifying targets even under strong interference. The reasons are as follows:
\subsubsection{Complex-Valued}
Current deep learning algorithms are primarily designed for optical real-valued images, hence they are unable to utilize the phase information of SAR images. The network proposed in this paper can directly manipulate complex-valued data within the network architecture. This not only preserves the complete information of the SAR images but also enhances the ability of model to capture the unique characteristics of SAR imagery.
\subsubsection{Fully Parameterized}
Unlike traditional neural networks, the network designed in this study does not employ dropout, downsampling, or similar operations but relies on the propagation of microwaves for interaction between layers. Thus, throughout the learning process of the network, no parameters are lost, which benefits the ability of network to recognize superimposed targets.
\subsubsection{Linear Modulation}
Electromagnetic wave propagation with dimensional independence facilitates unique linear modulation during spatiotemporal evolution. Consequently, this characteristic enables the EMWaveNet to demonstrate robust parallelism, particularly when overlapping SAR images. This parallelism not only ensures effective separation of overlaid images but also enhances the network's robustness against noise. By increasing the number of layers or neurons, the complexity of linear expressions within the network can be further enhanced, thereby improving its modulation capabilities.

\section{Conclusion}
Addressing the issue of current deep learning networks lacking in physical mechanisms and explainability in the SAR domain, this study introduces an explainable framework for SAR image recognition based on microwave propagation named EMWaveNet. The parameters learned within this framework possess distinct physical significance: they modulate the amplitude and phase of microwave propagation. The network inputs are complex-valued SAR images, utilizing the complex-valued information of SAR imagery to uncover latent physical features and enhance the physical logic behind decision-making. The independent propagation characteristics and the superposition properties of electromagnetic waves enable the network to possess de-overlapping capabilities and strong robustness. its robust performance offers a promising outlook for applications in complex SAR scenarios. Currently, EMWaveNet is presently in the early stages of development. Future research will aim to investigate additional applications of the network’s de-overlapping capabilities within the field of SAR.
\nocite{*}
\bibliographystyle{IEEEtran}
\bibliography{IEEEabrv,MWaveNet}

\end{document}